\documentclass[12pt]{article}

\usepackage{amssymb}
\usepackage{amsmath}
\usepackage[T1]{fontenc}
\usepackage{graphicx}
\usepackage{booktabs}
\usepackage{array}
\usepackage{caption}
\usepackage{multirow}
\usepackage{tabularx}
\usepackage{etoolbox}
\usepackage{makecell}
\usepackage{geometry}
\usepackage{tikz}
\usetikzlibrary{calc}
\usepackage{siunitx}
\usepackage{threeparttable}
\usepackage{subcaption}
\usepackage{comment}
\usepackage{longtable}
\usepackage{adjustbox}
\usepackage{float}
\usepackage{url}
\usepackage[breaklinks=true]{hyperref}
\usepackage[numbers,sort&compress]{natbib}

\begin{document}

\title{Satellite-based Rabi rice paddy field mapping in India: a case study on Telangana state}

\author{Prashanth Reddy Putta, Fabio Dell'Acqua\\
Dpt. of Electrical, Computer, Biomedical Engineering\\
University of Pavia\\
via Ferrata, 5, Pavia, IT-27100, PV, Italy}

\date{}

\maketitle

%% Abstract
\begin{abstract}
Accurate rice area monitoring is critical for food security and agricultural policy in smallholder farming regions, yet conventional remote sensing approaches struggle with the spatiotemporal heterogeneity characteristic of fragmented agricultural landscapes. This study developed a phenology-driven classification framework that systematically adapts to local agro-ecological variations across 32 districts in Telangana, India during the 2018-19 Rabi rice season. The research reveals significant spatiotemporal diversity, with phenological timing varying by up to 50 days between districts and field sizes ranging from 0.01 to 2.94 hectares. Our district-specific calibration approach achieved 93.3\% overall accuracy, an 8.0 percentage point improvement over conventional regional clustering methods, with strong validation against official government statistics (R² = 0.981) demonstrating excellent agreement between remotely sensed and ground truth data. The framework successfully mapped 732,345 hectares by adapting to agro-climatic variations, with Northern districts requiring extended land preparation phases (up to 55 days) while Southern districts showed compressed cultivation cycles. Field size analysis revealed accuracy declining 6.8 percentage points from medium to tiny fields, providing insights for operational monitoring in fragmented landscapes. These findings demonstrate that remote sensing frameworks must embrace rather than simplify landscape complexity, advancing region-specific agricultural monitoring approaches that maintain scientific rigor while serving practical policy and food security applications.

\end{abstract}


\vspace{0.5cm}

%% main text
\section{Introduction}

Rice cultivation stands at the nexus of global food security challenges, sustaining over half the world's population and providing up to 76\% of the caloric intake for Southeast Asian inhabitants~\cite{Rahman2023, Ahmad2022}. With global population projected to reach 8.5-10.6 billion by 2050~\cite{Valera2023}, the agricultural sector faces unprecedented pressure to increase production while managing resource constraints and climate uncertainties. Recent global rice market analyses indicate persistent volatility, with production reaching record levels of 543.3 million tonnes yet facing regional shortages and export restrictions that threaten food security~\cite{EssFeed2025}.

The urgency of this challenge has been amplified by recent geopolitical disruptions and climate impacts.~\cite{Valera2023} demonstrate that India's temporary export bans and climate-related production declines in Bangladesh and Sri Lanka illustrate the fragility of global rice supply chains. Such disruptions disproportionately affect import-dependent regions, particularly Sub-Saharan Africa and parts of Asia, where smallholder agricultural systems predominate and food security margins remain narrow. These smallholder systems, while contributing approximately 70\% of global food production, exhibit remarkable spatial and temporal variability that challenges standardized monitoring approaches~\cite{CuchoPadin2020, Getahun2024}.

In this context, accurate and timely crop monitoring becomes essential for multiple stakeholders: farmers require information for optimal management decisions, governments need data for policy formulation and food security planning, and international organizations require monitoring capabilities for early warning systems and humanitarian response~\cite{Wu2023}. However, smallholder agricultural systems present distinctive challenges for remote sensing applications, including small field sizes, variable planting and harvesting times, diverse management practices, mixed cropping patterns, and fuzzy field boundaries~\cite{CuchoPadin2020, Wang2022a}. These characteristics fundamentally alter the assumptions underlying many conventional crop mapping approaches.

The landscape of agricultural remote sensing has undergone substantial transformation over the past decade, driven by advances in satellite sensor capabilities, computational power, and algorithmic sophistication.~\cite{Joshi2023} identify four primary methodological classes in contemporary crop mapping research: spatial statistical methods, traditional machine learning approaches, phenology-based methods, and deep learning techniques. Traditional machine learning approaches, including Support Vector Machines, Random Forest, and Neural Networks, have demonstrated considerable success in crop mapping applications, achieving high accuracy when sufficient training data and computational resources are available~\cite{Victor2022}. Recent applications in crop improvement and sustainable production highlight the expanding role of machine learning across agricultural domains~\cite{Zhu2024}.

However, comprehensive reviews reveal that while deep learning approaches like CNN-LSTM combinations and multimodal fusion techniques show promise, they require substantial preprocessing, labeled datasets, and computational infrastructure~\cite{Bassine2023}. A critical gap exists between the sophisticated algorithms developed for uniform agricultural landscapes and their adaptability to heterogeneous smallholder systems.~\cite{Wu2023} emphasize that while machine learning and deep learning methods achieve impressive accuracy in crop monitoring, most struggle with the spatial and temporal variability characteristic of diverse agricultural landscapes, limiting their applicability across different agro-ecological contexts.

This adaptability gap has profound implications for understanding agricultural diversity across heterogeneous landscapes. Smallholder agricultural systems, which contribute approximately 70\% of global food production, exhibit remarkable spatial and temporal variability that challenges standardized monitoring approaches~\cite{CuchoPadin2020}. The diversity in field sizes, cropping patterns, irrigation practices, and phenological timing creates a complex mosaic that requires adaptive methodological frameworks rather than one-size-fits-all solutions~\cite{Getahun2024, Adegboye2021}.~\cite{Yuan2021} demonstrate that successful agricultural monitoring in diverse landscapes requires approaches that can accommodate local variations while maintaining methodological consistency across broader regions.

Despite substantial advances in agricultural remote sensing methodology, several critical gaps persist in understanding how monitoring approaches can adapt to landscape heterogeneity while maintaining operational viability. First, systematic evaluations of parameter calibration scale effects, comparing regional clustering versus administrative district-level optimization, remain limited in heterogeneous agricultural landscapes. Most studies either apply uniform parameters across entire study regions or acknowledge heterogeneity without developing systematic adaptation strategies. Second, while index-based methodologies offer distinct operational advantages, including computational efficiency, methodological transparency, and minimal training data requirements, systematic frameworks for adapting these approaches to diverse landscape conditions remain underexplored. Vegetation indices have maintained relevance in agricultural remote sensing due to their computational efficiency, interpretability, and reduced sensitivity to atmospheric effects~\cite{Ji2021}.

Phenology-based crop mapping represents a methodologically distinct approach that exploits the temporal dynamics of crop development cycles rather than relying on static spectral signatures.~\cite{Liu2021a} demonstrates the potential for phenology matching and cultivation pattern analysis using multi-source data integration, achieving robust rice mapping performance across diverse agricultural landscapes. Rice cultivation presents distinctive flooding, transplanting, and senescence phases that create identifiable spectral-temporal patterns. However,~\cite{wu2021} provide comprehensive assessments revealing that while various algorithms can correctly identify rice fields with simple cropping patterns, different products show low consistency in fragmented rice fields, and cloud prevalence in subtropical/tropical regions poses persistent challenges.

Current crop mapping literature reveals a methodological divide between algorithmic sophistication and landscape adaptability. Supervised machine learning models demonstrate high accuracy for crop identification in uniform landscapes, yet face challenges when applied across diverse agro-ecological conditions due to variations in data characteristics, phenological timing, and environmental heterogeneity. Deep learning approaches, while showing promise in controlled environments, often struggle with the transferability required for diverse landscape applications where training data may not be representative of the full range of local conditions~\cite{Victor2022}. This creates a methodological gap where the agricultural systems exhibiting the greatest landscape diversity are least well-served by standardized monitoring approaches.

The challenge of systematic parameter calibration across administrative boundaries represents a particularly critical research gap. The choice between regional clustering approaches, which offer theoretical efficiency and simplified implementation, and district-specific calibration frameworks, which provide enhanced adaptation to local conditions, has profound implications for both classification accuracy and operational feasibility. This methodological decision becomes particularly critical in resource-constrained environments where the balance between accuracy requirements and implementation complexity determines the practical utility of monitoring systems.

Telangana state exemplifies the complexities inherent in monitoring heterogeneous smallholder agricultural systems, where paddy cultivation defines the region's agricultural identity. The state's 33 districts encompass remarkable agro-ecological diversity, with field sizes averaging 0.16 hectares and ranging from fragmented plots to larger commercial operations. The Rabi rice season (December-May) represents a particularly strategic monitoring target, as unlike monsoon-dependent Kharif cultivation, Rabi production relies heavily on irrigation infrastructure, creating distinctive spectral-temporal patterns amenable to remote sensing analysis. However, dependency on diverse water sources introduces spatial variability in cultivation timing and management practices that complicates uniform monitoring approaches.

This study addresses the critical gap between methodological sophistication and landscape adaptability in agricultural remote sensing by developing and evaluating an adaptive phenological signature optimization framework for rice paddy mapping across heterogeneous smallholder systems. Our research pursues four interconnected objectives:

\textbf{(1)} Systematically characterize and optimize rice phenological signatures across 33 districts through district-specific temporal windows and adaptive parameter matrices, accommodating 67-day cultivation timing variations and heterogeneous field distributions.

\textbf{(2)} Develop and evaluate a hierarchical classification system incorporating selective temporal analytics (TSP/TPA), hybrid discrimination methods, and phenologically-aligned compositing strategies, with systematic evolution from regional clustering to district-specific calibration.

\textbf{(3)} Quantify the relationship between agricultural landscape heterogeneity and classification performance through field size stratification analysis, spatial fragmentation assessment, and phenological asynchrony impact evaluation across diverse agro-ecological contexts.

\textbf{(4)} Establish a robust validation framework integrating pixel-level accuracy, area estimation validation, and operational transferability evaluation using 953 manually digitized reference polygons across heterogeneous smallholder agricultural systems.

Our methodological framework addresses these gaps through systematic comparison of calibration strategies, comprehensive manual reference data development, and multi-dimensional validation protocols. The approach balances methodological sophistication with operational feasibility, demonstrating how systematic adaptation to landscape heterogeneity can enhance classification performance while maintaining computational efficiency suitable for resource-constrained operational environments. The hierarchical calibration framework provides insights into the relationship between agro-ecological diversity and optimal methodological parameters, offering a pathway for developing region-specific monitoring approaches that maintain scientific rigor while accommodating local environmental conditions. The findings have broader implications for agricultural monitoring in diverse landscapes globally, advancing understanding of how remote sensing methodologies can be systematically adapted to capture the complexity of heterogeneous agricultural systems while maintaining operational viability.

\section{Study Area}
\label{sec:study_area}

\subsection{Geographic Setting and Administrative Framework}
\label{subsec:geographic_setting}

This study covers 32 of 33 districts of the Telangana state in southern India (17.366°N, 78.476°E), covering approximately 112,077 km\textsuperscript{2} (Figure~\ref{fig:study_area}). Hyderabad district was excluded from the analysis due to its predominantly urban character and negligible agricultural land use. The study focuses on the Rabi rice season of 2018-19 (December 2018 to May/June 2019), during which irrigation-dependent cultivation dominates the agricultural landscape across the state's diverse agroecological zones.

\subsection{Agro-climatic Zonation and Environmental Characteristics}
\label{subsec:agroclimatic_zones}

The semi-arid tropical climate of Telangana presents distinct challenges to agricultural monitoring and necessitates a zone-based analytical approach. The state is administratively divided into three agroclimatic zones based on rainfall patterns, soil characteristics, and predominant cropping systems (Table~\ref{tab:agro-climatic-zones}), a classification framework extensively utilized in regional agricultural planning and research~\cite{PJTSAU2024}.

Rabi rice cultivation data for 2018-19 were obtained from two official sources: the Directorate of Economics and Statistics, Government of India~\cite{DES2019} and the Department of Agriculture, Government of Telangana~\cite{TelanganaAgri2019}. The \textbf{Northern Telangana Zone} encompasses ten districts including Nizamabad, Karimnagar, Jagtial, Peddapalli, Mancherial, and Nirmal, characterized by mixed irrigation infrastructure and cultivating 287,304 and 285,310 hectares of Rabi rice respectively according to the two data sources. The \textbf{Central Telangana Zone} covers eleven districts including Khammam, Siddipet, Jangoan, Warangal, and Hanumakonda, with extensive canal systems facilitating 173,334 and 163,535 hectares of rice production. The \textbf{Southern Telangana Zone} comprises twelve districts including Nalgonda, Suryapet, Yadadri Bhuvanagiri, Rangareddy, Vikarabad, and Mahabubnagar, combining canal and tank irrigation systems to support 281,870 and 250,959 hectares of Rabi rice cultivation\cite{CRIDA2024, TelanganaStats2024}.

\subsection{Landscape Heterogeneity and Agricultural Context}
\label{subsec:landscape_heterogeneity}

The environmental variability between these agroclimatic zones (Table~\ref{tab:agro-climatic-zones}) significantly influences agricultural practices and creates distinct spectral-temporal signatures relevant to satellite-based monitoring. Soil characteristics vary markedly across the state, with black cotton soils (Vertisols) predominating in certain regions and red lateritic soils (Alfisols) dominating others, creating zone-specific agricultural management practices that affect crop phenology and spectral response patterns.

Approximately 60\% of the state's agricultural area operates under rainfed conditions~\cite{pingle2011}, making crop production highly sensitive to monsoon variability and creating temporal uncertainty in cultivation patterns. The agricultural landscape is characterized by smallholder farming systems, with average landholding size of 1.12 hectares in 2010-11, slightly below the national average of 1.16 hectares~\cite{HansIndia2017}. This fragmented agricultural structure presents unique challenges for satellite-based crop area estimation, as field sizes often approach or fall below the spatial resolution of freely available satellite imagery, necessitating sub-pixel analysis techniques and careful consideration of mixed-pixel effects in remote sensing applications.

The irrigation infrastructure varies significantly across zones, with the northern and central zones benefiting from extensive canal networks fed by the Krishna and Godavari river systems, while the southern zone relies on a combination of canal irrigation and traditional tank systems. This heterogeneous irrigation landscape creates diverse crop calendars and varying degrees of agricultural intensification, requiring zone-specific remote sensing approaches for accurate crop area assessment.

\begin{table}
\centering
\caption{Characteristics of agro-climatic zones in Telangana}
\label{tab:agro-climatic-zones}
\resizebox{\columnwidth}{!}{%
\begin{tabular}{cccccc}
\hline
\textbf{\begin{tabular}[c]{@{}c@{}}Agro-\\ climatic \\ Zone\end{tabular}} &
  \textbf{\begin{tabular}[c]{@{}c@{}}Annual \\ Rainfall \\ (mm)\end{tabular}} &
  \textbf{\begin{tabular}[c]{@{}c@{}}Major \\ Soil Types\end{tabular}} &
  \textbf{\begin{tabular}[c]{@{}c@{}}Irrigation \\ (\% area)\end{tabular}} &
  \textbf{\begin{tabular}[c]{@{}c@{}}Dominant \\ Cropping Pattern\end{tabular}} &
  \textbf{\begin{tabular}[c]{@{}c@{}}Avg. \\ Field Size (ha)\end{tabular}} \\
\hline
\begin{tabular}[c]{@{}c@{}}Northern   \\ Telangana\end{tabular} &
  900-1100\textsuperscript{a} &
  \begin{tabular}[c]{@{}c@{}}Black (50\%),\\ Red (35\%),\\ others (15\%)\textsuperscript{b}\end{tabular} &
  35-40\textsuperscript{c} &
  \begin{tabular}[c]{@{}c@{}}Rice-Maize/\\ Redgram/Cotton\textsuperscript{d}\end{tabular} &
  1.0-1.5\textsuperscript{e} \\
\hline
\begin{tabular}[c]{@{}c@{}}Central\\ Telangana\end{tabular} &
  779-1213\textsuperscript{a} &
  \begin{tabular}[c]{@{}c@{}}Red (54\%),\\ Calcareous(13\%),\\ Black (6\%)\textsuperscript{b}\end{tabular} &
  45-50\textsuperscript{c} &
  \begin{tabular}[c]{@{}c@{}}Cotton-Rice/\\ Maize/Greengram\textsuperscript{d}\end{tabular} &
  0.8-1.2\textsuperscript{e} \\
\hline
\begin{tabular}[c]{@{}c@{}}Southern\\ Telangana\end{tabular} &
  606-853\textsuperscript{a} &
  \begin{tabular}[c]{@{}c@{}}Red (65\%),\\ Black (20\%),\\ others (15\%)\textsuperscript{b}\end{tabular} &
  25-30\textsuperscript{c} &
  \begin{tabular}[c]{@{}c@{}}Rice-Groundnut/\\ Sorghum/Cotton\textsuperscript{d}\end{tabular} &
  0.5-1.0\textsuperscript{e} \\
\hline
\end{tabular}%
}
\vspace{0.2cm}
\footnotesize
\begin{tabular}{@{}l@{}}
\textsuperscript{a}Professor Jayashankar Telangana State Agricultural University (2024); \\
\textsuperscript{b}Government of Telangana Planning Department (2024); \\
\textsuperscript{c}Pingle (2011); \\
\textsuperscript{d}Central Research Institute for Dryland Agriculture (CRIDA) (2024); \\
\textsuperscript{e}The Hans India (2017) \\
\end{tabular}
\end{table}

\subsection{Administrative Boundary Management}
\label{subsec:boundary_management}

To address the challenge of changing administrative boundaries in our time-series analysis, we standardized all district references to the current 33-district framework of Telangana (Government of Telangana, 2024) \cite{TelanganaStats2024}. We implemented a comprehensive district name mapping dictionary that resolved variations between our remote sensing boundaries and official agricultural statistics from both the Directorate of Economics and Statistics (2019) \cite{DES2019} and the Telangana Department of Agriculture (2019)\cite{TelanganaAgri2019}.

The mapping system resolved three key challenges: variant spellings (e.g., ``Jagitial''/``Jagtiyal''), administrative renaming (e.g., ``Warangal Urban'' to ``Hanumakonda''), and partial identifiers (e.g., ``Medchal'' to ``Medchal Malkajgiri''). By maintaining a consistent spatial reference framework throughout our analysis, we mitigated a significant potential source of error in agricultural monitoring across administrative reorganizations.

\begin{figure}[!htbp]
    \centering
    \includegraphics[width=\textwidth]{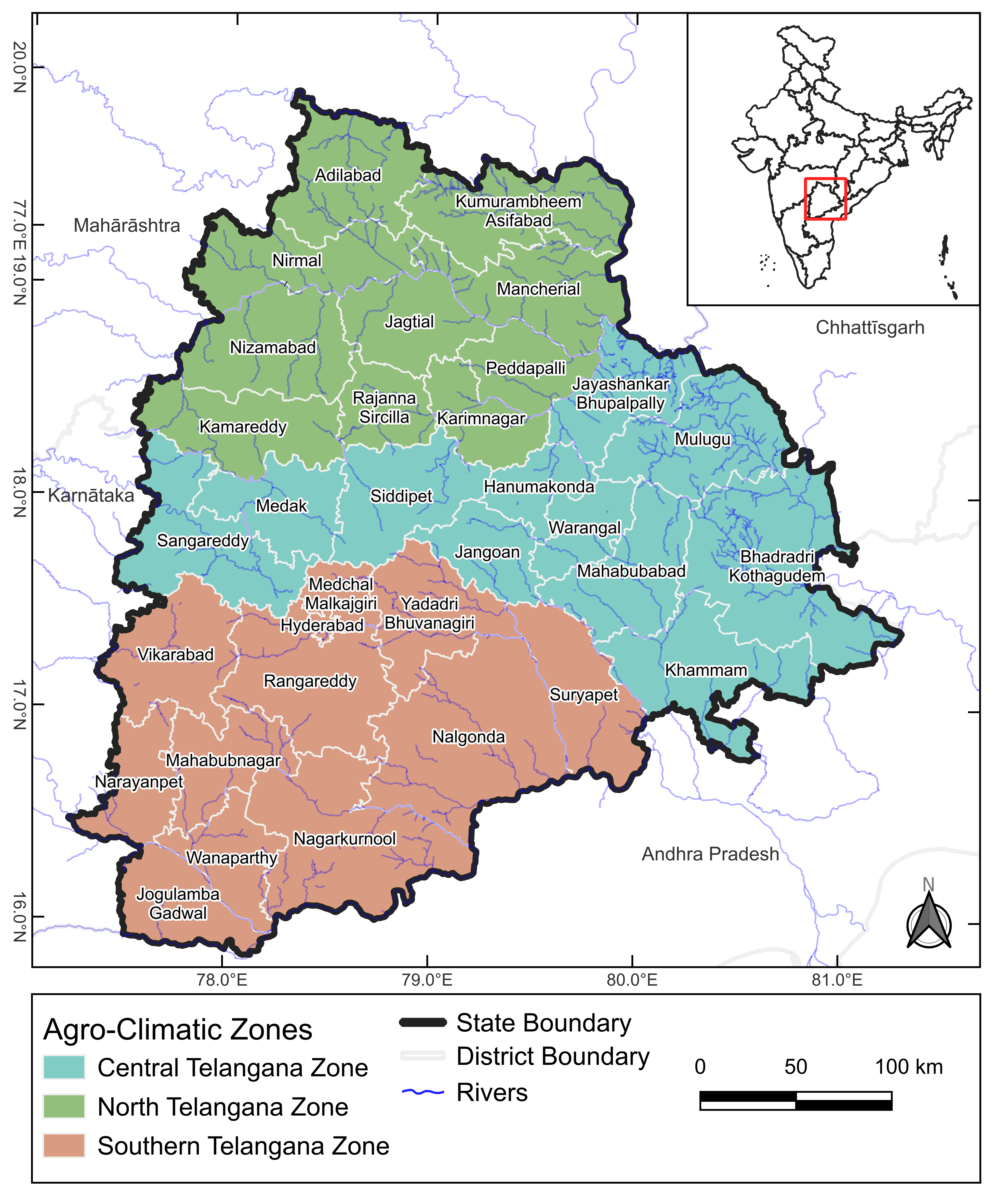}
    \caption{Telangana State, India, showing 33 districts across three agro-climatic zones (Northern, Central, and Southern), with major rivers and neighboring states. Inset shows Telangana's location within India.}
    \label{fig:study_area}
\end{figure}

\section{Methodological Approach: Regional Clustering versus District-Level Parameter Calibration}
\label{sec:methodological_approach}
\subsection{Rationale for Scale Selection in Agricultural Remote Sensing}
\label{subsec:scale_rationale}
A fundamental consideration in remote sensing-based crop classification concerns the optimal spatial scale for parameter calibration~\cite{Lu2023, Wang2019}. In regions with diverse agricultural landscapes such as Telangana, where paddy cultivation practices vary substantially across agro-ecological zones, this decision significantly impacts classification accuracy and operational feasibility~\cite{Bellon2017, Foley2005}.

\subsection{Initial Regional Clustering Strategy}
\label{subsec:clustering_strategy}
We initially pursued a regional clustering approach, grouping Telangana's 33 districts into four agro-ecological clusters based on environmental variables and agricultural patterns~\cite{Bellon2017, Delegido2011}: (1) \textbf{Cluster 1} (8 districts), (2) \textbf{Cluster 2} (5 districts), (3) \textbf{Cluster 3} (4 districts), and (4) \textbf{Cluster 4} (16 districts). This approach offered theoretical advantages of computational efficiency and simplified parameter management while maintaining regional agricultural relevance~\cite{McRoberts2007, Tornos2015}.

\subsection{Limitations of the Clustering Approach}
\label{subsec:clustering_limitations}
Preliminary validation revealed critical limitations that compromised classification effectiveness. \textbf{Phenological asynchrony} within clusters showed temporal variations up to 10 days between districts, significantly affecting spectral signatures during key growth stages~\cite{Gao2021, Liu2022}. \textbf{Parameter divergence} was substantial, with optimal spectral threshold values varying by more than 20\% between districts within the same cluster~\cite{Cao2019, Gitelson2004}, for example, optimal NDVI/LSWI ratio ranges differed from 1.8--2.5 to 1.8--3.0 within supposedly homogeneous groups. \textbf{Inconsistent performance} within clusters was particularly pronounced in complex landscapes, where districts required extensive customization that negated the efficiency benefits of clustering~\cite{Chen2013, Xiao2005}.
Most critically, several districts within each cluster required such substantial parameter modifications that the fundamental assumption of within-cluster homogeneity was invalidated. This observation suggested that the spatial scale of administrative districts better captured the relevant agricultural and environmental variability for accurate parameter calibration~\cite{McRoberts2007}.

\subsection{Methodological Evolution to District-Level Approach}
\label{subsec:district_evolution}
Based on these limitations, we evolved our methodology to implement district-specific parameter calibration. This approach maintains the same underlying classification framework but acknowledges that local variations in soil conditions, water management practices, planting schedules, and field characteristics require finer-scale parameterization than regional clustering can accommodate~\cite{Potgieter2021, Yang2019}.
The district-level approach incorporates locally calibrated spectral thresholds, district-specific phenological windows, and adaptive temporal analytics while maintaining computational feasibility across the study region. This methodological choice prioritizes classification accuracy and reliability over the theoretical efficiency of regional clustering, recognizing that operational crop monitoring systems require consistent performance across diverse agricultural landscapes~\cite{Jin2023, Kussul2016}.

\subsection{Framework for Scale Selection}
\label{subsec:scale_framework}
This experience suggests that researchers should evaluate several criteria when selecting calibration scales~\cite{Atkinson2012, Woodcock1987}: phenological timing variations within proposed units ($>$7--10 days indicating need for finer scales), parameter optimization divergence between units ($>$15--20\% threshold variations), accuracy requirements for the intended application, landscape fragmentation characteristics, and the degree of customization required to achieve acceptable performance. When preliminary analysis indicates extensive customization needs, direct implementation at finer administrative scales may prove more efficient than hierarchical clustering approaches~\cite{Friedl2010, Defourny2019}.

\section{Materials and Methods}
\label{sec:materials_methods}

\begin{figure}[!htbp]
    \centering
    \includegraphics[width=0.78\textwidth]{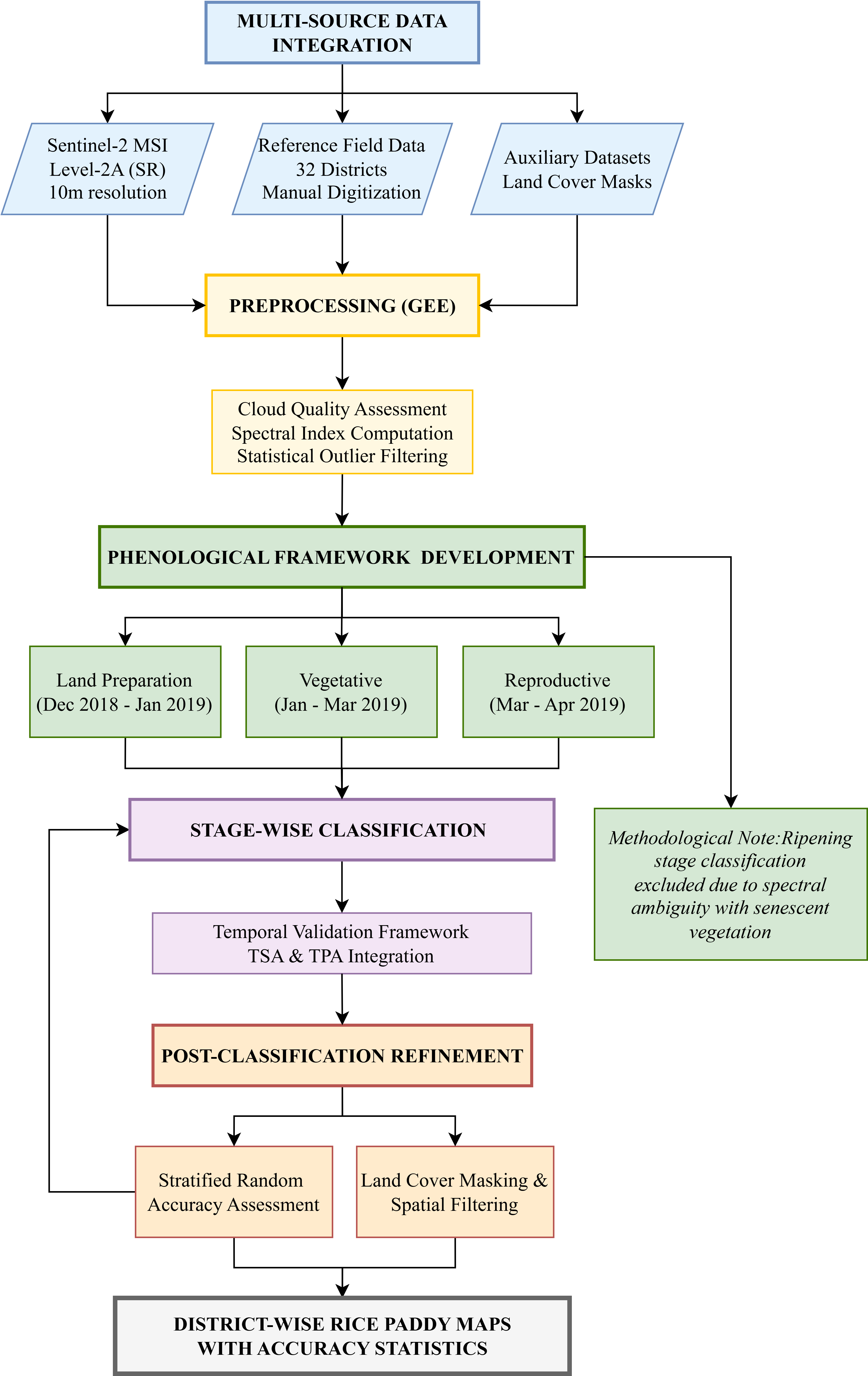}
    \caption{Rice paddy mapping framework for Telangana State using multi-temporal Sentinel-2 data with phenological classification and temporal validation across 32 districts.}
    \label{fig:methodological_framework}
\end{figure}

Our approach employs a phenology-driven classification framework that adapts to Telangana's heterogeneous agricultural landscapes. Figure~\ref{fig:methodological_framework} illustrates the complete workflow, which integrates multi-temporal Sentinel-2 imagery, spectral-temporal analysis, and district-specific parameter optimization to achieve robust rice paddy mapping across diverse agro-climatic zones.

\subsection{Study Design and Temporal Framework}
\label{subsec:study_design}

The analysis covered 32 of 33 districts in Telangana state, focusing on the 2018-19 Rabi rice season (December 2018 to May / June 2019), when irrigation-dependent cultivation dominates. Hyderabad district was excluded due to its predominantly urban character, resulting in a total study area of approximately 112,077 km².

\textbf{Temporal Analysis Framework:} The study period captures the complete Rabi rice phenological cycle across Telangana's diverse agro-climatic zones (Table~\ref{tab:temporal_windows}). District-specific growing seasons were defined based on local agricultural calendars, with transplanting typically occurring from December to January and harvest extending from April to June, depending on district-specific cultivation practices.

\begin{table}[!htbp]
\centering
\caption{District-specific temporal analysis windows for Rabi rice classification}
\label{tab:temporal_windows}
\begin{tabular}{lccc}
\toprule
\makecell{\textbf{Agro-climatic} \\ \textbf{Zone}} & \textbf{Transplanting} & \textbf{Peak Growth} & \textbf{Harvest} \\
\midrule
\makecell{Northern Telangana \\ (10 districts)} & Dec 15 - Jan 15 & Feb 15 - Mar 15 & May 1 - Jun 15 \\
\makecell{Central Telangana \\ (11 districts)} & Dec 1 - Jan 10 & Feb 1 - Mar 10 & Apr 15 - May 30 \\
\makecell{Southern Telangana \\ (12 districts)} & Dec 20 - Jan 20 & Feb 20 - Mar 20 & May 10 - Jun 20 \\
\bottomrule
\end{tabular}
    \begin{tablenotes}
        \footnotesize
        \item \textit{Note:} December dates refer to 2018; January dates and beyond refer to 2019.
    \end{tablenotes}
\end{table}

\textbf{District-Specific Calibration Rationale:} Based on preliminary cluster analysis showing substantial intra-cluster parameter variation (>20\% threshold divergence and >10-day phenological differences), district-level parameter calibration was adopted to accommodate local variations in cultivation practices, field size distributions, and irrigation management systems across Telangana's heterogeneous agricultural landscape.

\subsection{Data Sources}
\label{subsec:data_sources}

\subsubsection{Satellite Imagery}
\label{subsubsec:satellite_imagery}

We used Sentinel-2 multispectral imagery (10 m resolution, Harmonized Sentinel-2 MSI, Level-2A (SR) collection) acquired through Google Earth Engine as the primary data source. The Sentinel-2 constellation provides high spatial resolution (10-20m) combined with frequent revisit capability ($\approx$5 days), making it well-suited for capturing rice phenology dynamics. The imagery dataset spanned the 2018--19 Rabi season with district-specific temporal windows to encompass complete rice cultivation cycles. Based on previous regional phenological studies~\cite{Boschetti2017, Dong2016}, we established cultivation period windows for each district: the standard window (applied to most districts) extended from mid-December 2018 to late May 2019; two northern districts with later planting patterns (Mancherial, Kumurambheem Asifabad) used a February 1, 2019--June 20, 2019 window; and several other districts had slightly adjusted end dates based on local harvesting patterns. These temporally-tailored acquisition windows ensured comprehensive coverage of all phenological stages across the region's diverse cultivation timelines.

\subsubsection{Auxiliary Data}
Supplementary datasets included JRC Global Surface Water masks (permanent and seasonal)~\cite{Pekel2016,Pekel2017}, ESA WorldCover 10m land cover~\cite{Zanaga2021}, and administrative boundaries from Telangana Open Data Portal~\cite{TelanganaOpenData2024}.

\subsection{Reference Data Preparation}
\label{subsec:reference_data}

Reference data were manually digitized using a custom Google Earth Engine annotation tool, creating 953 polygons (460 non-paddy, 493 paddy) across all districts. Sampling intensity was proportional to district rice production importance, ranging from 10 to 90 polygons per district.

These reference polygons served as the basis for both training and internal validation during the threshold calibration process. Only later, during the final accuracy assessment (see Section~\ref{subsec:accuracy_assessment}), was a stratified random sampling strategy applied to ensure statistically sound performance evaluation across field sizes and land cover types.

The significant variation in reference polygon counts across districts (ranging from 5 to 60 samples per class) was deliberately implemented to reflect both the importance of rice cultivation in each district and our confidence in identifying representative fields. Major rice-producing districts with significant contributions to state production (e.g., Nalgonda with 60 non-paddy and 30 paddy polygons) received more intensive sampling to ensure robust algorithm training and validation. Conversely, districts with marginal rice cultivation or less complex agricultural landscapes (e.g., Adilabad with 5 polygons per class) received proportionally fewer samples. This stratified approach to reference data collection balanced the need for comprehensive representation against practical field sampling constraints, while ensuring sufficient data points to characterize the spectral-temporal signatures across diverse agro-ecological contexts.

\subsection{Rice Phenological Framework}
\label{subsec:rice_phenological}

\subsubsection{Growth Stage Definition}
\label{subsubsec:growth_stage}

Building on established rice phenology models~\cite{Boschetti2017, Dong2016, Gorelick2017}, we defined four sequential growth stages:

\begin{enumerate}
\item \textbf{Land Preparation/Initial Establishment} -- characterized by flooded, puddled fields with minimal vegetation.
\item \textbf{Early Vegetative} -- rapid leaf emergence and tillering, with exponential leaf area increase.
\item \textbf{Reproductive} -- panicle initiation, heading and flowering, corresponding to maximum canopy greenness.
\item \textbf{Ripening} -- grain filling and senescence, marked by declining greenness.
\end{enumerate}

Stage transitions were determined through analysis of NDVI temporal profiles from reference fields, using smoothed trajectories (Savitzky-Golay filter: window=7, order=3) to identify when >50\% of fields crossed stage-specific thresholds.

\subsubsection{Spectral Indices Calculation}
\label{subsubsec:spectral_indices}

Five spectral indices (Table~\ref{tab:spectral_indices})  were calculated to capture vegetation and water dynamics:

\begin{table}[!ht]
    \centering
    \renewcommand{\arraystretch}{1.4}
    \caption{Spectral indices used for rice paddy identification and monitoring}
    \label{tab:spectral_indices}
    \begin{threeparttable}
    \begin{tabular}{
        >{\raggedright\arraybackslash}p{5.2cm} 
        >{\centering\arraybackslash}m{6.3cm} 
        >{\raggedright\arraybackslash}p{3.2cm}
    }
        \toprule
        \textbf{Index Name} & \textbf{Formula} & \textbf{Reference} \\
        \midrule
        Normalized Difference Vegetation Index (NDVI) 
        & $\displaystyle \frac{B8 - B4}{B8 + B4}$ 
        & Tucker (1979) \\

        Modified Normalized Difference Water Index (MNDWI) 
        & $\displaystyle \frac{B3 - B11}{B3 + B11}$ 
        & Xu (2006) \\

        Land Surface Water Index (LSWI) 
        & $\displaystyle \frac{B8 - B11}{B8 + B11}$ 
        & Xiao et al. (2006) \\

        Enhanced Vegetation Index (EVI) 
        & $\displaystyle 2.5 \times \frac{B8 - B4}{B8 + 6B4 - 7.5B2 + 1}$ 
        & Huete et al. (2002) \\

        Soil-Adjusted Vegetation Index (SAVI) 
        & $\displaystyle 1.5 \times \frac{B8 - B4}{B8 + B4 + 0.5}$ 
        & Huete (1988) \\
        \bottomrule
    \end{tabular}
    \begin{tablenotes}
        \footnotesize
        \item \textit{Note:} B2 = Blue, B3 = Green, B4 = Red, B8 = Near-Infrared (NIR), B11 = Short-Wave Infrared (SWIR) bands from Sentinel-2 imagery.
    \end{tablenotes}
    \end{threeparttable}
\end{table}

These indices were chosen for their known relevance to rice phenology: NDVI and EVI track canopy greenness, LSWI/MNDWI capture water presence, and SAVI moderates soil background. Using published formulations ensures consistency with prior studies.

\subsection{Data Processing}
\label{subsec:data_processing}

\subsubsection{Preprocessing}
\label{subsubsec:preprocessing}

The Sentinel-2 imagery required substantial preprocessing to ensure data quality and temporal consistency across the study period. We implemented a comprehensive preprocessing workflow in Google Earth Engine that addressed both atmospheric and sensor-related artifacts while preserving the temporal signal necessary for phenological analysis.

To mitigate cloud interference, which poses a significant challenge for optical remote sensing in tropical regions, we applied a two-step filtering approach:

\begin{enumerate}
\item \textbf{Cloud Masking}: We first excluded entire scenes with cloud coverage exceeding 80\% of the district area, ensuring baseline image quality. For remaining scenes, we applied pixel-level cloud and cirrus masking using the Sentinel-2 QA60 band, which provides reliable cloud detection based on radiometric properties. This implementation followed established approaches in the Google Earth Engine platform~\cite{Gorelick2017} and is particularly effective for tropical agricultural monitoring where intermittent cloudiness can otherwise compromise time-series analysis.

\item \textbf{Statistical Outlier Removal}: For cases where the QA60 band failed to identify all anomalies (particularly cloud shadows and thin clouds), we applied a secondary statistical filtering approach based on the temporal behavior of each spectral index. Specifically, we identified and masked pixels whose values deviated beyond 2$\times$ the interquartile range (IQR) in either direction from the median of their respective time-series. This bi-directional filtering approach effectively removed both anomalously high values (typically associated with undetected clouds or cloud edges) and anomalously low values (typically associated with cloud shadows or sensor errors) while preserving the natural biological variability in rice phenology signals. The IQR-based approach was selected over simple standard deviation thresholding to accommodate the non-normal distribution of index values commonly observed in agricultural time series. This method effectively removed anomalous values while preserving the natural variability of rice phenology signals.

\item \textbf{Radiometric normalization}: After cloud and outlier masking, we applied radiometric normalization by scaling all reflectance values to a consistent 0--1 range while preserving temporal metadata. This standardization facilitates meaningful comparison of spectral indices across the growing season.
\end{enumerate}

This linear rescaling process preserves the relative spectral relationships while standardizing the value range across all images, thereby eliminating the effects of varying illumination conditions and facilitating consistent threshold application throughout the phenological cycle. The normalized reflectance values were then used for all subsequent spectral index calculations and analyses.

For the reference data analysis, which was conducted separately from the main image processing stream, we exported temporal spectral profiles from our manually digitized polygons (described in Section~\ref{subsec:reference_data}) to CSV format for detailed analysis in Python. During this reference data analysis phase, we applied a Savitzky--Golay smoothing filter (window=7, order=3) to reduce noise and fill minor gaps in the temporal signatures of known paddy and non-paddy areas. This reference data processing informed our threshold selection and parameterization but was distinct from the main image preprocessing workflow in GEE.

\subsubsection{Resampling and Compositing}
\label{subsubsec:resampling_compositing}

The multi-resolution nature of Sentinel-2 bands (10m, 20m, and 60m native resolutions) presents challenges for integrated analysis~\cite{Kowaleczko2023}. Rather than explicitly resampling all bands to a common resolution, which can introduce artifacts and information loss, we implemented a more nuanced approach that preserves the native resolution characteristics of each band during index calculation.

We allowed Google Earth Engine to handle mixed-resolution indices natively during computation~\cite{Gorelick2017}, following a conservative approach that prevents artificial upsampling of coarser bands. For example, when computing the Modified Normalized Difference Water Index (MNDWI), which combines the 10m green band (B3) with the 20m short-wave infrared band (B11), the system implicitly aligns them to the coarsest resolution (20m) during the calculation to preserve data fidelity. This approach maintains the integrity of the spectral information while avoiding the introduction of artificial detail that might compromise classification accuracy. Final outputs (indices and classification maps) were exported at 10m resolution to maximize spatial detail for agricultural fields, which are frequently smaller than 1 hectare in the study region.

Unlike conventional approaches that create temporal composites at regular calendar intervals (e.g., monthly or 16-day composites)~\cite{Holben1986}, we constructed composites aligned with rice phenological stages. Specifically, we averaged all available cloud-free observations within each of the four identified rice growth stages (Land Preparation, Early Vegetative, Reproductive, and Ripening) to produce one representative ``stage-composite'' per district. This phenologically-driven compositing strategy yields more biologically meaningful signals for classification than arbitrary time intervals~\cite{Meraner2020}, as it accounts for the district-specific variations in planting dates and growth rates. By aligning image composites with actual crop development stages rather than calendar dates, we minimized confusion between different growth phases that might otherwise occur in fixed-interval composites, particularly in regions with asynchronous planting practices.

This phenologically-aligned compositing approach represents a methodological advancement over fixed-calendar compositing techniques that cannot adapt to regional and seasonal variations in crop development. By creating stage-specific composites rather than date-specific ones, we effectively normalized the temporal dimension according to biological growth stages rather than calendar dates. This approach mitigates the problem of mixed phenological signals that can occur when rice fields at different growth stages are included in the same temporal composite due to asynchronous planting dates, which is particularly common in smallholder agricultural landscapes~\cite{Mungai2016} where planting decisions are influenced by diverse factors including water availability, labor constraints, and local agricultural practices.

\subsection{Classification Methodology}
\label{subsec:classification_methodology}

\subsubsection{Parameter Optimization}
\label{subsubsec:parameter_optimization}

We employed a phenology-driven paddy classification, multi-index threshold approach~\cite{Xiao2005, Dong2016, Boschetti2017}. We developed a Python-based workflow to statistically evaluate threshold ranges for each index and stage. For each index-stage combination, we tested three candidate percentile ranges: 10--90th (broad), 25--75th (interquartile), and a custom literature-based range~\cite{Sakamoto2019}. We then selected the range achieving the best balance of sensitivity and specificity on the reference fields. In practice, the broadest (10--90th) range was often optimal, providing robust coverage of field variability. This hybrid biophysical-statistical tuning ensures that thresholds align with both agronomic insight and reference data statistics.

\subsubsection{Stage-Specific Detection Criteria}
\label{subsubsec:stage_specific_criteria}

Based on the optimization analysis, we developed tailored spectral thresholds and criteria for each growth stage to address the distinct phenological characteristics of paddy rice~\cite{Son2021}:

\begin{itemize}
\item \textbf{Land Preparation Stage:} For flooded paddies with sparse vegetation, we applied district-specific approaches based on local optimization. In 29 districts, we implemented ratio-based criteria including NDVI/SAVI ratios (typically $<$ 2.30-2.70), SAVI/NDVI ratios ($>$ 0.35-0.50), and specialized conditions such as MNDWI-NDVI differences and EVI/LSWI ratios for complex landscapes. For comparison, we also evaluated the LSWI-EVI relationship method, which identifies paddy fields when LSWI $\geq$ (EVI - 0.05). We complemented the selected approach with modest NDVI thresholds (0.15-0.30) across districts to account for minimal vegetation cover~\cite{Li2013}. Additionally, LSWI thresholds (0.10-0.45) and positive MNDWI values ($>$ 0) were employed to enhance detection of standing water while excluding permanent water bodies.

\item \textbf{Vegetative Stage:} During rapid canopy development, we implemented broader NDVI ranges (0.25-0.70) and EVI thresholds ($\geq$ 0.15-0.45) to capture increasing greenness~\cite{Zheng2016}. LSWI values between 0.20-0.50 were applied to ensure detection of adequate soil moisture. In districts requiring enhanced discrimination, we applied spectral ratio criteria including NDVI/EVI $<$ 2.0 AND LSWI/EVI $>$ 0.5-0.7, or NDVI/LSWI ratios (1.2-3.0). Many districts achieved satisfactory discrimination using basic thresholds without additional ratio constraints, following our principle of algorithm simplicity where performance was adequate.

\item \textbf{Reproductive Stage:} During dense canopy development, we employed elevated NDVI thresholds (0.45-0.95) and higher EVI values (0.25-0.70) across districts. In 15 districts requiring enhanced discrimination, we implemented NDVI/LSWI ratio criteria with ranges of approximately 1.35-3.5, often combined with additional constraints such as EVI/LSWI $<$ 1.8-2.2 and NDVI/EVI $<$ 1.6-1.8. The remaining 17 districts achieved excellent discrimination using basic spectral thresholds without additional ratio constraints, demonstrating effective performance with simpler approaches.

\item \textbf{Ripening Stage:} As chlorophyll content declines, we employed wider NDVI windows (0.15-0.70) and lower MNDWI values ($<$ -0.35) to capture senescence patterns. In districts requiring enhanced discrimination, we applied SAVI/NDVI ratios (typically $>$ 0.60-0.68) and NDVI/LSWI ratios ($>$ 3.0) or specialized SAVI/NDVI ranges (0.55-1.30)~\cite{Huete1988}. However, due to persistent discrimination challenges with other vegetation types during senescence, ripening stage areas were ultimately \textbf{excluded} from the final area calculations.
\end{itemize}

\subsubsection{Spectral Ratio Criteria}
\label{subsubsec:spectral_ratio_criteria}

We developed spectral index ratio criteria to enhance paddy discrimination across all growth stages in 32 districts of Telangana~\cite{Maxwell2018}. We systematically evaluated multiple discrimination approaches including: (1) ratio-based combinations (NDVI/LSWI, EVI/LSWI, NDVI/EVI, NDVI/SAVI, SAVI/NDVI, MNDWI-NDVI differences) derived from reference data distributions~\cite{Bannari1995}, and (2) the LSWI-EVI relationship method~\cite{Xiao2006}, which identifies paddy fields when LSWI $\geq$ (EVI - 0.05).

For each approach, we calculated discrimination effectiveness using our reference polygons across all growth stages and districts. Multi-condition approaches combining 2-3 ratio constraints were tested for enhanced discrimination during complex phenological stages~\cite{Teluguntla2018}. Following the principle of algorithm simplicity, ratio-based criteria were implemented only in districts where basic spectral thresholds proved insufficient for adequate discrimination. This approach maintained computational efficiency while ensuring optimal classification performance across diverse landscape conditions. Based on systematic evaluation (Section~\ref{subsec:spectral_performance}), we implemented a hybrid approach using district-specific optimal methods determined through empirical validation.

\subsubsection{Temporal Analytics Integration}
\label{subsubsec:temporal_analytics}

We refined spectral classification using temporal consistency checks through two complementary approaches: \textbf{Temporal Stability Parameters (TSP)} and \textbf{Temporal Pattern Analysis (TPA)}, both calibrated with district-specific thresholds to account for regional variations in rice phenology. We strategically selected NDVI as the sole index for temporal stability analysis based on substantial scientific evidence supporting its reliability in crop monitoring across the entire growth cycle~\cite{Huete2002}.

\paragraph{Temporal Stability Parameters (TSP)}
For each growth stage and district, we calculated NDVI standard deviation within the stage period for known paddies. Unlike conventional approaches that apply uniform thresholds across entire study regions~\cite{Zhang2018}, we established district-specific and phenological stage-specific maximum allowable NDVI standard deviation thresholds. This exploited the fact that managed paddy fields typically exhibit relatively uniform phenology within each growth phase compared to other land covers~\cite{Li2012}. Pixels exhibiting temporal variability exceeding these thresholds were excluded from classification as likely mixed or non-paddy areas.

\paragraph{Temporal Pattern Analysis (TPA)}
To balance classification accuracy with computational efficiency, we selectively applied TPA in districts where paddy identification was particularly challenging due to spectral confusion with other land covers. In these districts, we established four district-specific thresholds: (1) Peak NDVI minimum and maximum values, (2) NDVI increase between early and peak stages, and (3) NDVI decrease from peak to late stages. These thresholds were calibrated using ground-truth data from each district, accounting for local variations in rice phenology~\cite{Li2012}. Districts with simpler land cover patterns relied solely on TSP and traditional spectral classification to maintain computational efficiency.

Detailed threshold values and implementation results for both approaches are presented in Section~\ref{subsec:temporal_enhancement}.

\subsection{Classification Refinement Approaches}
\label{subsec:classification_refinement}

\subsubsection{Statistical Outlier Filtering}
\label{subsubsec:outlier_filtering}

To remove anomalous pixels, we applied statistical filtering on the composite index images. Using the interquartile range (IQR) method, we masked any pixel whose index value deviated beyond 2$\times$IQR from the median~\cite{Boschetti2017}. This was done separately for NDVI, EVI, LSWI composites at each stage. In practice, this filtering reduced noisy outlier pixels caused by residual clouds or shadows. For example, in Karimnagar district tailoring the IQR multiplier per index (2$\times$ for NDVI/EVI, 1.5$\times$ for LSWI) cut commission errors by $\sim$14\% and omission by $\sim$17\%. Across districts, applying this outlier check (especially in the Reproductive stage) consistently improved accuracy (e.g. Warangal's accuracy rose from 76\% to 85.8\% after filtering).

\subsubsection{Land Cover Integration}
\label{subsubsec:land_cover_integration}

In many cases we used external land cover maps to exclude impossible areas. We used the ESA WorldCover 10 m product~\cite{Zanaga2021} to mask out classes that are not cropland: trees (forest), shrublands, grasslands, built-up, and bare/sparse cover. Removing these pixels (which could otherwise mimic paddy spectral patterns) further reduced commission errors. For example, eliminating woody areas or villages prevented spurious detections in regions like Jayashankar Bhupalpally and Kumurambheem Asifabad. We quantified this effect: in Jogulamba Gadwal, combining water masking and land-cover filtering reduced mapped area by 18.9\%, correcting many false positives on reservoirs and grasslands.

\subsubsection{Spatial Refinement}
\label{subsubsec:spatial_refinement}

As a final processing step, we implemented spatial coherence refinement on the paddy classification to enhance the integrity of field boundaries. Our approach converts all intermediate classification results to explicit binary values (0/1), ensuring data consistency throughout the analysis workflow.

After completing spectral classifications and exclusion operations, we applied a focal-mode filter with a 20-meter radius to the combined binary paddy mask. This spatial filtering technique employs a moving window majority rule to eliminate isolated pixels and fill small gaps in field boundaries~\cite{Lu2007}. The filter effectively addresses salt-and-pepper noise typical in pixel-based classification while preserving the essential characteristics of field geometries.

The spatial refinement significantly improved boundary delineation and visual coherence of the paddy fields, reflecting the natural contiguity of agricultural landscapes. This post-processing step is particularly important for paddy rice mapping, as the distinctive field structure of rice cultivation systems requires clear boundary definition for accurate area estimation and pattern analysis~\cite{Dong2016}.

\subsection{Classification Workflow and Area Calculation}
\label{subsec:classification_workflow}

The classification workflow integrated the previously described steps as illustrated in Figure~\ref{fig:methodological_framework}, ensuring consistent methodology while accommodating district-specific threshold variations. This workflow leveraged the distinctive spectral and temporal signatures, while implementing the analytical approaches demonstrated in Figure~\ref{fig:ndvi_trajectories_nalgonda}.

One critical decision was to \textbf{exclude the Ripening stage} from final area tallies. In practice, the Ripening masks vastly overestimated area (due to confusion with bare or senescent vegetation). For example, in Jogulamba Gadwal the raw Ripening-stage mask covered 58,501 ha vs. only 8,083 ha identified in earlier stages. Similar vast overestimates occurred in several districts. Including Ripening-stage results would have introduced severe commission errors consistent with findings by~\cite{Boschetti2017} and our own. Therefore, the final rice area was based on Land Preparation through Reproductive stages only.

\subsection{Accuracy Assessment and Validation}
\label{subsec:accuracy_assessment}

\subsubsection{Validation Approach}
\label{subsubsec:validation_approach}

We implemented a dual validation approach to assess both classification accuracy and area estimation performance, addressing two fundamental aspects of agricultural mapping validation: pixel-level classification performance and landscape-scale area estimation accuracy.

\textbf{Pixel-Level Classification Validation:} Given the significant disparity in field sizes between paddy and non-paddy land cover classes in Telangana, we adopted an area-based stratified random sampling strategy for classification accuracy assessment. Our analysis revealed that 79.3\% of paddy fields are smaller than 0.2 hectares, and 93.4\% are below 0.5 hectares, with a median size of only 0.091 hectares. In contrast, non-paddy land cover classes exhibit greater size heterogeneity, ranging from 0.028 to 94.662 hectares, with 23.1% exceeding 2 hectares.

Without size-based stratification, conventional random sampling would systematically underrepresent the smallholder paddy fields that dominate the agricultural landscape but occupy proportionally less spatial area. To address this sampling bias, we categorized fields into four strata: tiny (<0.5 ac), small (0.5–2.0 ac), medium (2.0–10.0 ac), and large (>10.0 ac). Validation points were allocated proportionally across size categories, with enhanced sampling weight for paddy fields to reflect their economic importance despite smaller spatial extent.

Classification performance was evaluated using standard accuracy metrics including overall accuracy, balanced accuracy, F1-score, Kappa coefficient, and producer/user accuracies for both classes, calculated separately for each district and size category.

\textbf{Area Estimation Validation:} To validate district-level area estimates, we compared our mapped rice areas against two independent official datasets: (1) Government of India Directorate of Economics and Statistics (2018-2019)\cite{DES2019}, and (2) Telangana Department of Agriculture (2019)~\cite{TelanganaAgri2019}. This dual-reference approach provides robustness against potential inconsistencies in official statistics and enables assessment of mapping performance relative to different data collection methodologies.
Area validation employed comprehensive statistical metrics including coefficient of determination (R²), Root Mean Square Error (RMSE), Mean Absolute Error (MAE), bias, Pearson correlation coefficient (r), Bias, and linear regression parameters. These metrics collectively assess correlation strength, prediction accuracy, systematic error tendencies, and agreement with reference data.
This comprehensive validation framework ensures statistical validity across the diverse land-use and parcel-size spectrum while providing both pixel-level and landscape-scale assessment capabilities, aligning with best practices in agricultural validation studies~\cite{Boschetti2017, Dong2016}.

\subsection{Implementation Platform}
\label{subsec:implementation}

All image processing was conducted in Google Earth Engine, with statistical analysis and visualization in Python 3.8 and QGIS. Custom tools included the Advanced Land Cover Mapping Tool for reference data collection and specialized time-series analysis scripts.

\section{Results}
\label{sec:results}

\subsection{Spectral Phenological Characterization}
\label{subsec:spectral_phenology}

\begin{figure}[!htbp]
    \centering
    \includegraphics[width=\textwidth]{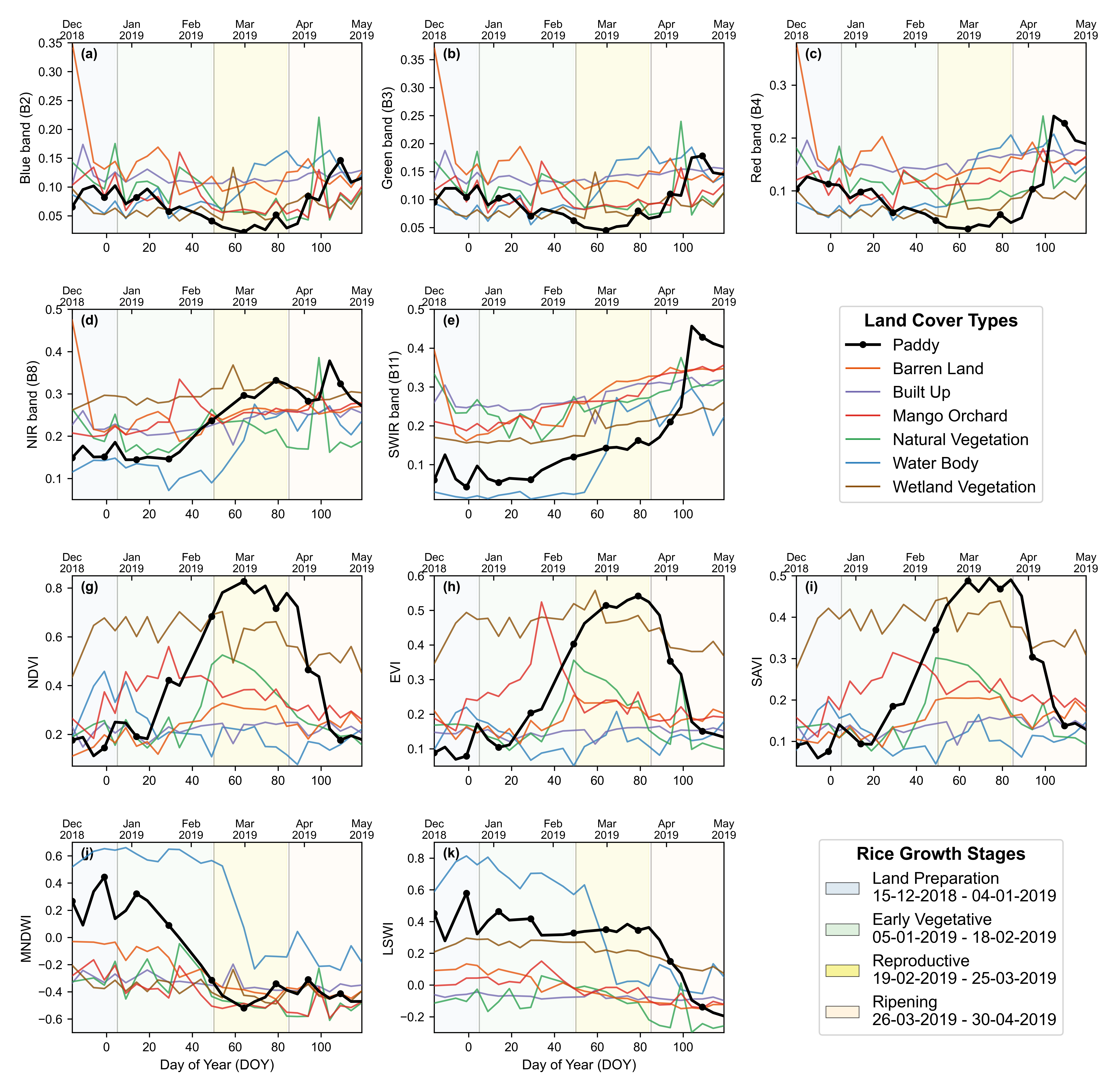}
    \caption{Temporal dynamics of spectral indices averaged over randomly selected sample fields for each land cover type in Nalgonda district, highlighting rice’s distinctive phenological trajectory through four growth stages. Shaded regions indicate stage boundaries.}
    \label{fig:results_spectral}
\end{figure}

The temporal spectral curves reveal three characteristics that distinguish rice paddies and justify our stage-wise discrimination:

1. \textbf{Flooding signature in land preparation.} During the shaded land-preparation window, rice’s mean MNDWI and LSWI (Fig.~\ref{fig:results_spectral}j,k) rise to approximately 0.4–0.6, whereas other cover types remain near zero or negative. This pronounced moisture signal underpins the use of the LSWI–EVI relationship in this stage, as flooded soil in rice fields creates a spectral environment where that rule is highly effective.

2. \textbf{Peak biomass in reproductive stage.} In the reproductive window, rice’s mean NDVI (Fig.~\ref{fig:results_spectral}g) reaches ~0.75–0.85, with EVI (Fig.~\ref{fig:results_spectral}h) ~0.45 and SAVI (Fig.~\ref{fig:results_spectral}i) ~0.5, substantially higher than other land covers. The clear separation of rice curves from others in this stage corresponds to the high discrimination accuracy reported later in the paper, confirming that spectral ratio or threshold-based criteria are most powerful here.

3. \textbf{Smooth, consistent trajectories within stages.} The aggregated rice curves (e.g., NDVI in Fig.~\ref{fig:results_spectral}g) show gradual, predictable increases and declines between stages, in contrast to irregular or flat patterns of natural vegetation, built-up areas, or water bodies. This temporal stability, reflecting synchronized agricultural management, motivates our temporal stability parameters, which flag deviations inconsistent with the expected phenological progression.

Taken together, rice exhibits a canonical four-stage spectral-temporal trajectory: low values during land preparation (flooded soil), steady increase in vegetative growth, sustained peak in reproduction, and controlled decline in ripening that differs fundamentally from other covers that remain relatively constant or fluctuate irregularly. These aggregated patterns provide the empirical basis for establishing district-specific thresholds that accommodate local variations, but rely on the underlying temporal signature to distinguish rice from spectrally similar but phenologically different vegetation types.

\subsection{Temporal Dynamics of Spectral Indices Across Districts}
\label{subsec:temporal_dynamics}

Our analysis revealed significant district-specific variations in spectral-temporal signatures across Telangana's diverse agro-climatic zones. Figure~\ref{fig:indices_districts} presents the temporal dynamics of key spectral indices across six representative districts, demonstrating the phenological and spectral diversity found across all 32 districts: Nirmal and Kumurambheem Asifabad (Northern Telangana Zone), Warangal and Khammam (Central Telangana Zone), and Suryapet and Nalgonda (Southern Telangana Zone).

\subsubsection{District-Specific Spectral Variations}

The six representative districts exhibit markedly different spectral index patterns, illustrating the complexity found across Telangana's agricultural landscape. Nirmal (Figure~\ref{fig:indices_districts_a}) shows highly variable NDVI patterns with multiple fluctuations during vegetative growth, while Kumurambheem Asifabad (Figure~\ref{fig:indices_districts_b}) displays smoother but delayed spectral development with later peak timing.

Central zone districts demonstrate contrasting behaviors: Warangal (Figure~\ref{fig:indices_districts_c}) exhibits gradual, steady NDVI progression with well-defined stage transitions, whereas Khammam (Figure~\ref{fig:indices_districts_d}) shows rapid spectral increases and higher amplitude variations during peak growth periods.

Southern zone districts reveal distinct early-season patterns: Nalgonda (Figure~\ref{fig:indices_districts_e}) demonstrates compressed temporal signatures with early peak development, while Suryapet (Figure~\ref{fig:indices_districts_f}) shows the most consistent and smooth spectral progression across all phenological stages.

LSWI patterns vary significantly between districts, with clear flooding signals in northern districts contrasting with more subdued water index responses in southern irrigated areas. EVI and SAVI indices similarly show district-specific amplitude and timing variations that reflect local growing conditions and management practices~\citep{Xiao2005,Huete2002}.

\subsubsection{Implications for Classification}

These temporal variations in spectral signatures across districts highlight the complexity of paddy classification in Telangana's diverse agricultural landscape. The observed differences in phenological timing and spectral magnitude demonstrate why district-specific calibration was essential for accurate classification. The findings underscore that a uniform classification approach would inadequately capture the state's agricultural diversity, supporting our decision to implement district-specific threshold optimization and stage-specific discrimination criteria.

% Page 1: Subfigures 4a and 4b
\begin{figure}[H]
    \centering
    % First subfigure
    \begin{subfigure}[b]{\linewidth}
        \centering
        \includegraphics[height=0.45\textheight]{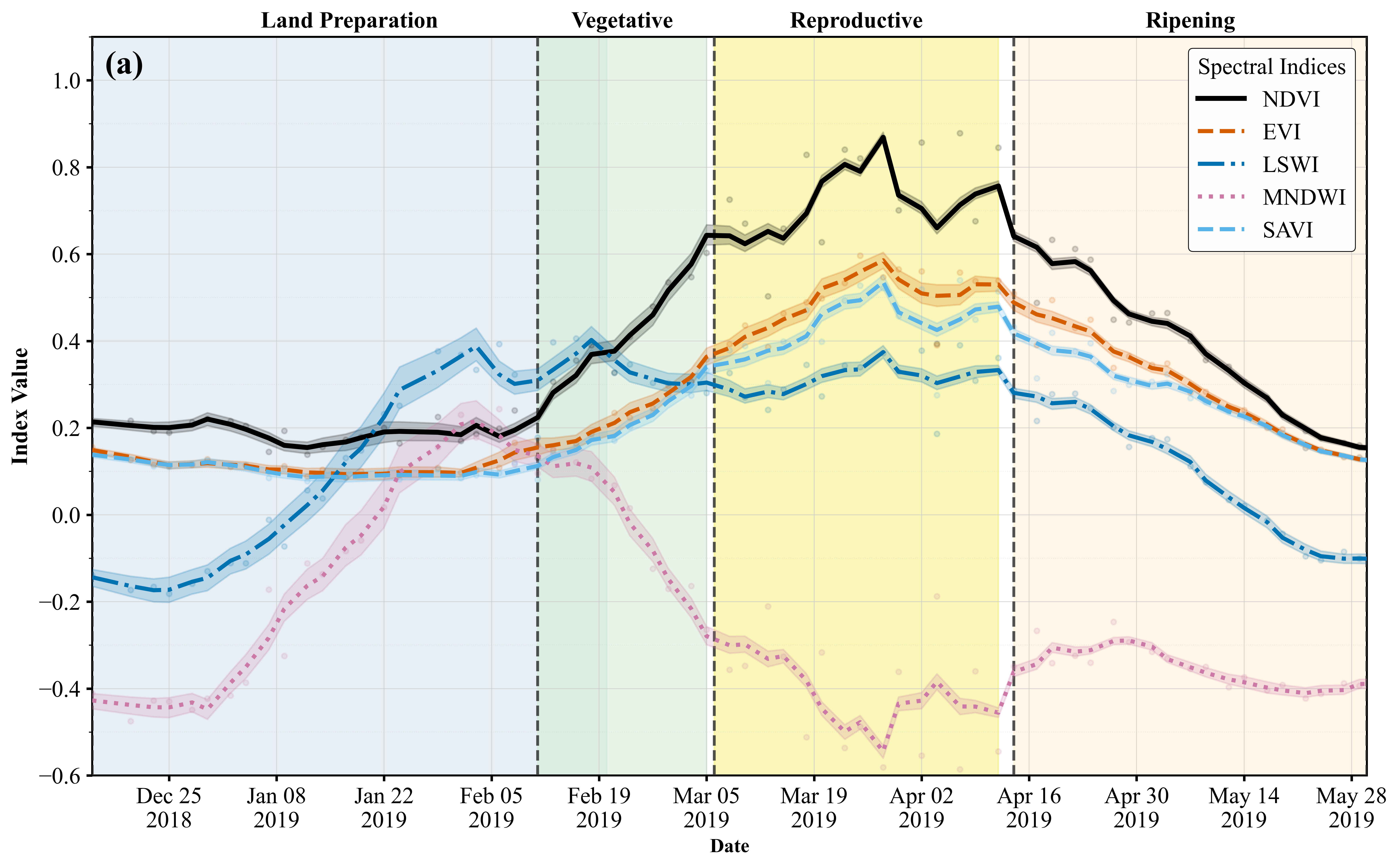}
        \caption{Nirmal District}
        \label{fig:indices_districts_a}
    \end{subfigure}
    
    %\vspace{0.5cm}
    
    % Second subfigure
    \begin{subfigure}[b]{\linewidth}
        \centering
        \includegraphics[height=0.45\textheight]{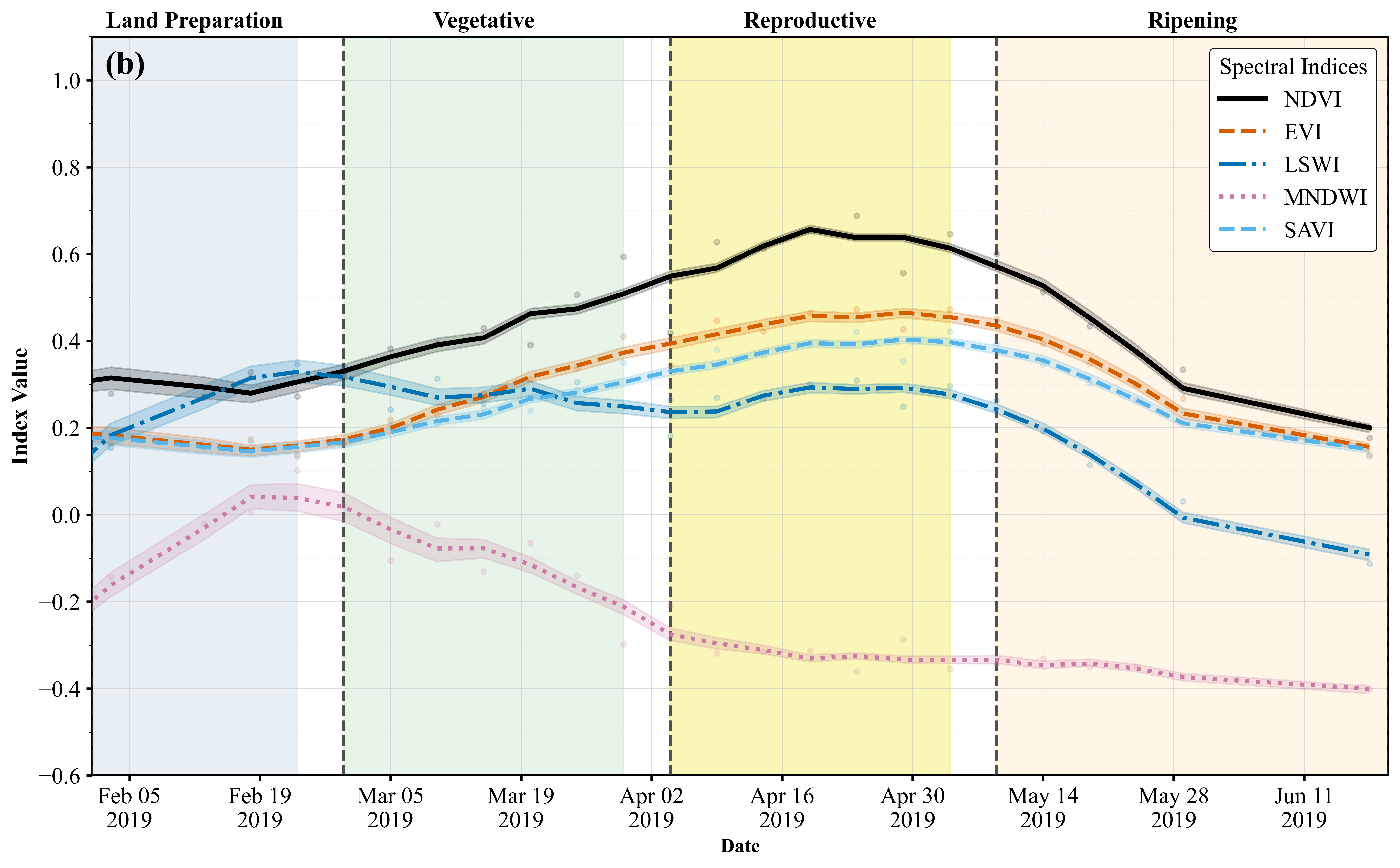}
        \caption{Kumurambheem Asifabad District}
        \label{fig:indices_districts_b}
    \end{subfigure}

    \caption[]{Figure~\ref{fig:indices_districts} continued.}
    \label{fig:indices_districts}
\end{figure}

%\clearpage

% Page 2: Subfigures 4c and 4d (continued)
\begin{figure}[H]
    \ContinuedFloat
    \centering
    
    % Third subfigure
    \begin{subfigure}[b]{\linewidth}
        \centering
        \includegraphics[height=0.45\textheight]{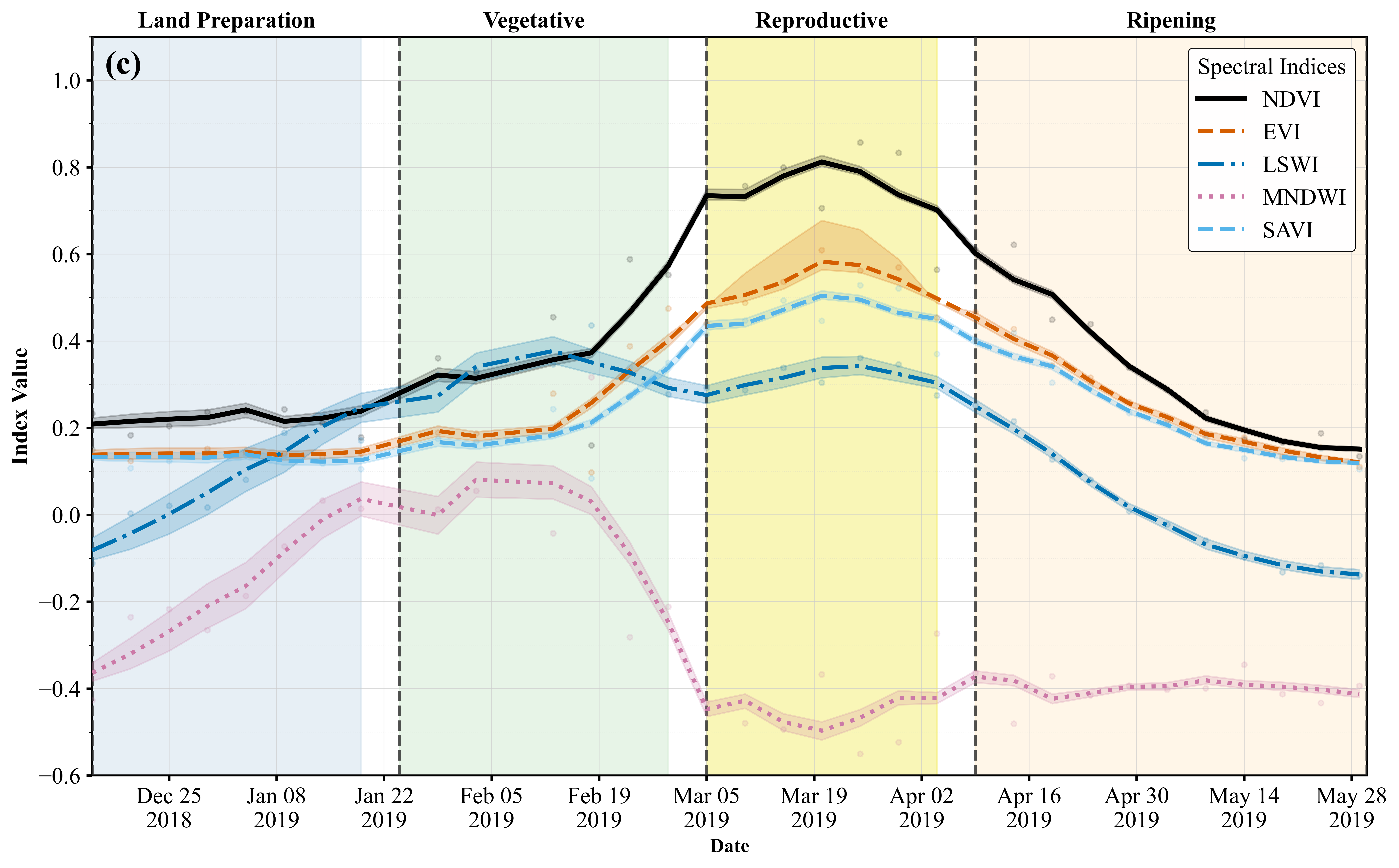}
        \caption{Warangal District}
        \label{fig:indices_districts_c}
    \end{subfigure}
    
    %\vspace{0.5cm}
    
    % Fourth subfigure
    \begin{subfigure}[b]{\linewidth}
        \centering
        \includegraphics[height=0.45\textheight]{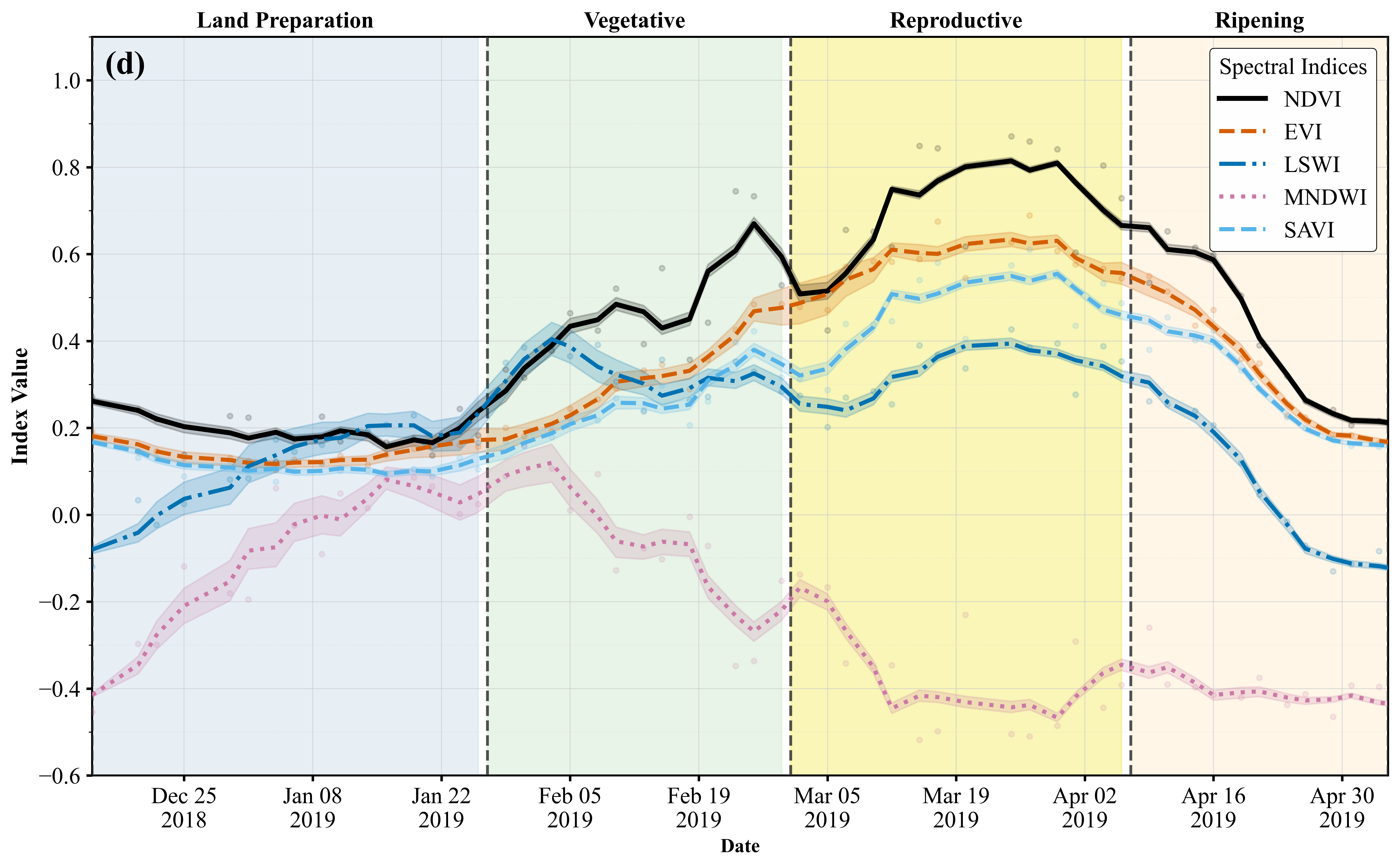}
        \caption{Khammam District}
        \label{fig:indices_districts_d}
    \end{subfigure}
    
    \caption[]{Figure~\ref{fig:indices_districts} continued.}
\end{figure}

%\clearpage

% Page 3: Subfigures 4e and 4f (continued)
\begin{figure}[H]
    \ContinuedFloat
    \centering
    
    % Fifth subfigure
    \begin{subfigure}[b]{\linewidth}
        \centering
        \includegraphics[height=0.45\textheight]{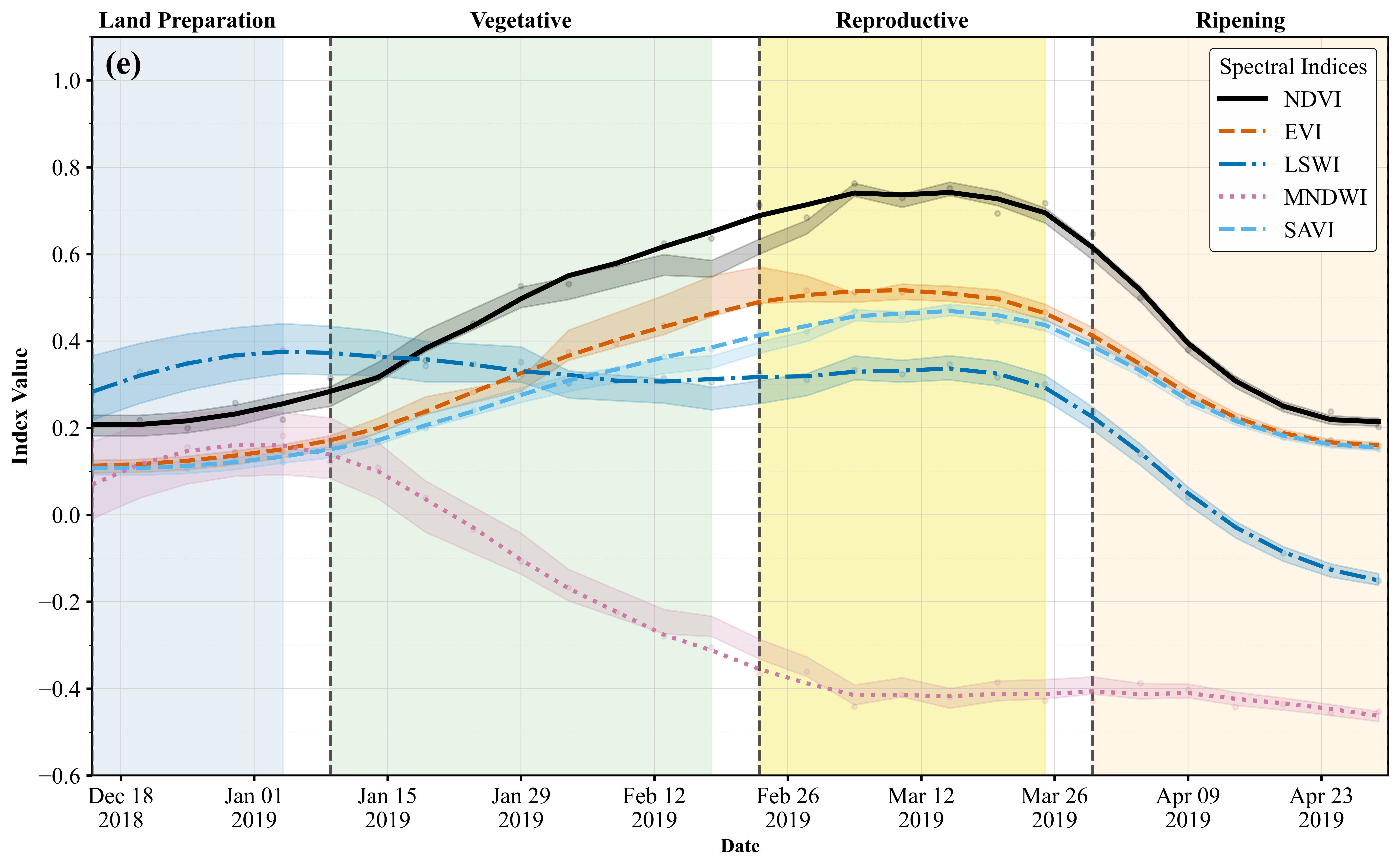}
        \caption{Nalgonda District}
        \label{fig:indices_districts_e}
    \end{subfigure}
    
    %\vspace{0.5cm}
    
    % Sixth subfigure
    \begin{subfigure}[b]{\linewidth}
        \centering
        \includegraphics[height=0.45\textheight]{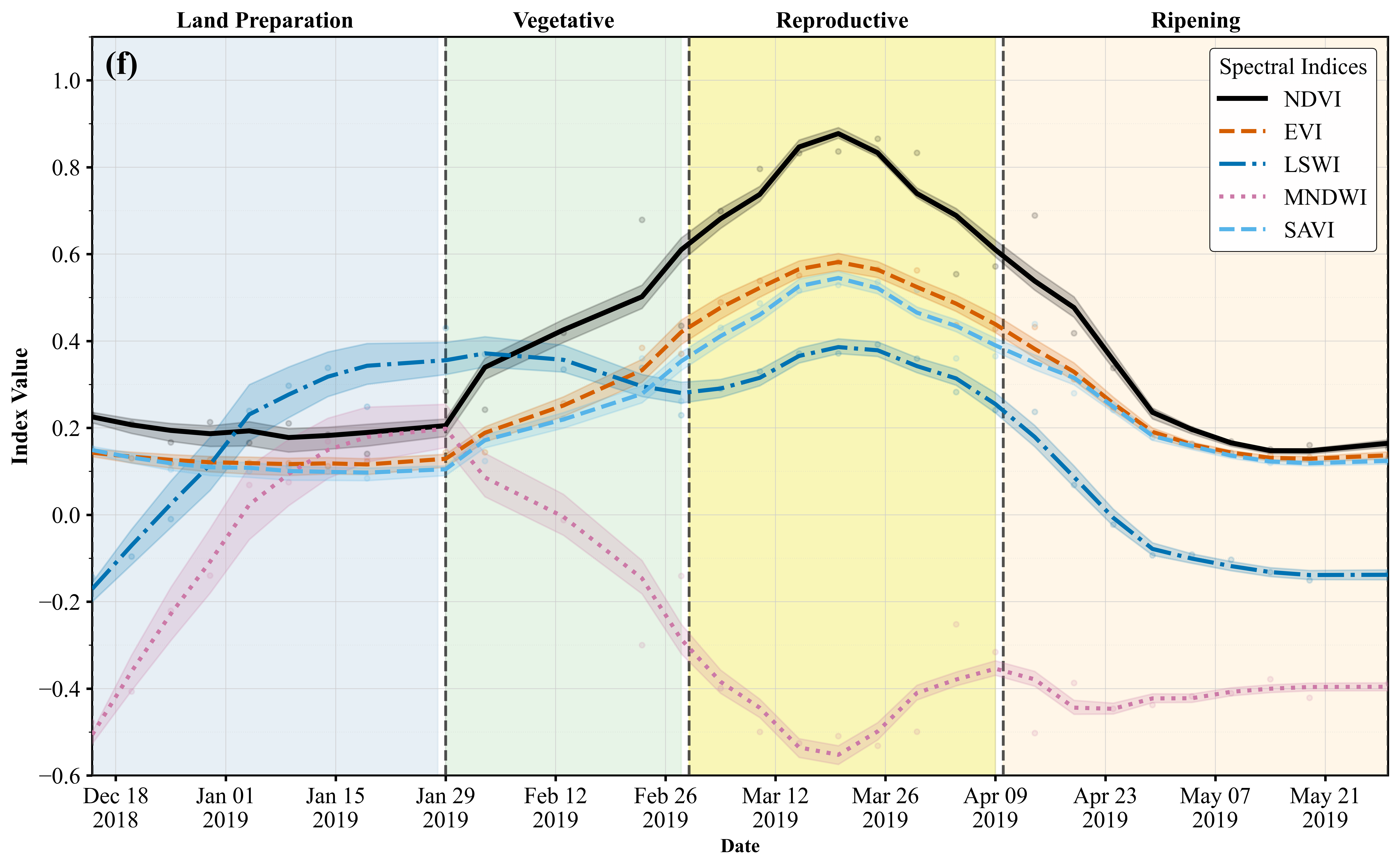}
        \caption{Suryapet District}
        \label{fig:indices_districts_f}
    \end{subfigure}

    \caption{Spectral indices temporal profiles during rice growth: (a) Nirmal, (b) Kumurambheem Asifabad, (c) Warangal, (d) Khammam, (e) Nalgonda, (f) Suryapet. Phenological boundaries (dashed).}
\end{figure}

\subsection{Phenological Signatures Across Telangana's Rice Agro-ecologies}
\label{subsec:phenological_signatures}

Analysis of rice phenological patterns revealed remarkable spatiotemporal heterogeneity across Telangana's agricultural landscape, with cultivation timing varying by up to 50 days between districts. This heterogeneity reflects fundamental adaptations to water availability, soil characteristics, and cropping system constraints across the state's three agro-climatic zones (Northern, Central, and Southern).

\begin{figure}[!htbp]
    \centering
    \includegraphics[width=\textwidth]{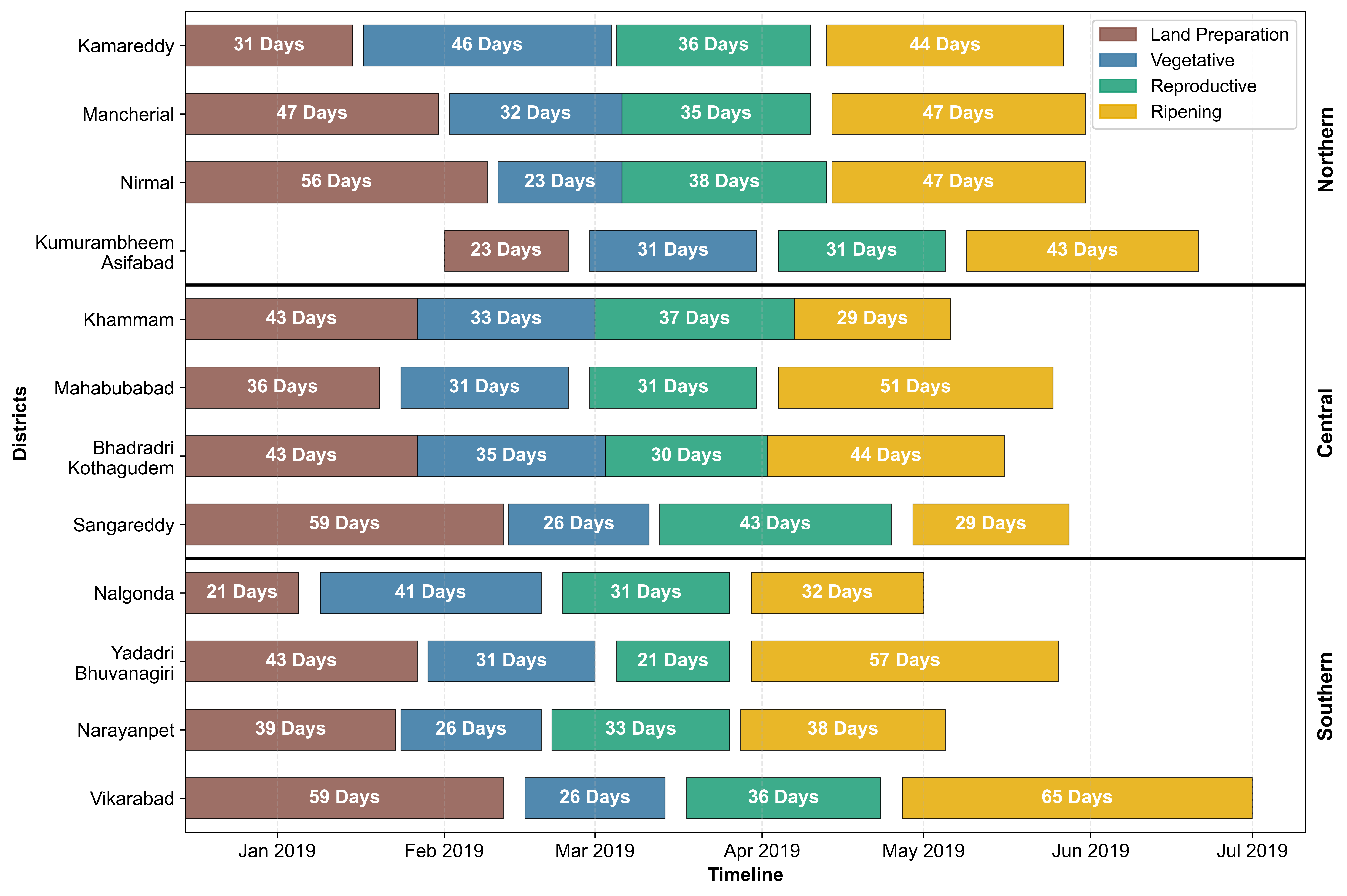}
    \caption{Temporal progression of rice phenology across three agro-climatic zones of Telangana, showing stage durations for 12 representative districts. Districts are grouped by zone (Northern, Central, and Southern) and ordered by planting date within each zone.}
    \label{fig:results_phenology}
\end{figure}

\begin{table}[!htbp]
\centering
\caption{Rice phenological metrics across Telangana's agro-climatic zones based on analysis of 32 districts}
\label{tab:phenology_comprehensive}
\begin{tabular}{lcccc}
\toprule
\textbf{Metric} & \textbf{Northern} & \textbf{Central} & \textbf{Southern} & \textbf{State Range} \\
\midrule
\multicolumn{5}{l}{\textit{Planting dates}} \\
Earliest & Jan 14 & Jan 20 & Jan 9 & Jan 9 \\
Latest & Feb 28 & Feb 13 & Feb 16 & Feb 28 \\
Range (days) & 45 & 24 & 38 & 50 \\
\midrule
\multicolumn{5}{l}{\textit{Stage durations (days, mean ± SD)}} \\
Land preparation & 35 ± 9 & 39 ± 7 & 37 ± 10 & 20-58 \\
Vegetative & 32 ± 9 & 33 ± 4 & 30 ± 6 & 17-50 \\
Reproductive & 40 ± 6 & 34 ± 4 & 34 ± 6 & 20-48 \\
Ripening & 46 ± 2 & 45 ± 9 & 46 ± 11 & 28-64 \\
Total duration & 153 ± 10 & 151 ± 6 & 148 ± 15 & 121-182 \\
\midrule
Districts (n) & 10 & 11 & 11 & 32 \\
\bottomrule
\end{tabular}
\end{table}

The most pronounced pattern was the north-south gradient in cultivation timing (Figure~\ref{fig:results_phenology} and Table~\ref{tab:phenology_comprehensive}). Northern districts initiated land preparation 5-50 days later than southern districts, with Kumurambheem Asifabad beginning on February 28, compared to January 9 in Nalgonda.

Stage duration variability provided additional insights into regional adaptation strategies. Notable extremes included Nirmal's extended land preparation phase (55 days) and Vikarabad's prolonged ripening period (64 days). The vegetative stage ranged from 17 days (Jogulamba Gadwal) to 50 days (Adilabad), while the reproductive stage varied from 20 days (Yadadri Bhuvanagiri) to 48 days (Rajanna Sircilla). These variations likely reflect cultivar selection and management practices adapted to local constraints. Complete district-wise phenological data is provided in Supplementary Figure~\ref{fig:district_phenology}.

\subsection{Spectral Index Discrimination Performance}
\label{subsec:spectral_performance}

Our systematic evaluation revealed district-specific optimal discrimination methods across Telangana's diverse agricultural landscape. The analysis demonstrated that optimal discrimination approaches varied substantially by both growth stage and local conditions, necessitating a strategic approach balancing performance with algorithm complexity.

For the Land Preparation stage, systematic evaluation across 29 districts with ratio-based criteria revealed superior average effectiveness (71.2\%) compared to LSWI-EVI methods (65.9\%). Ratio-based approaches proved more effective in 15 districts (51.7\%), while LSWI-EVI achieved better results in 11 districts (37.9\%), with comparable performance in the remaining districts. Performance variations reflected differences in local hydrological conditions, field management practices, and landscape heterogeneity. Districts achieving excellent performance with ratio-based methods ($>$90\% effectiveness) included Hanumakonda, Karimnagar, Nalgonda, and Suryapet, while LSWI-EVI showed competitive performance in districts with specific water management characteristics. Three districts (Adilabad, Kumurambheem Asifabad, and Medchal Malkajgiri) relied solely on basic spectral thresholds due to adequate discrimination without additional constraints.

During the Vegetative stage, analysis revealed that many districts achieved satisfactory discrimination using basic NDVI, EVI, and LSWI thresholds without requiring additional ratio constraints. In districts where enhanced discrimination was necessary, both ratio-based methods (NDVI/LSWI ratios, NDVI/EVI combinations) and LSWI-EVI relationships demonstrated effective performance. Districts implementing ratio criteria included those with complex agricultural landscapes or mixed cropping patterns, while simpler agricultural systems maintained high accuracy with basic thresholds. This pattern supported our strategic decision to apply ratio constraints selectively rather than universally.

The Reproductive stage exhibited exceptional discrimination performance, with 15 districts implementing NDVI/LSWI ratio criteria achieving consistently high effectiveness. Districts applying ratio constraints included Hanumakonda, Nalgonda, Nizamabad (with multi-condition approaches achieving near-perfect discrimination), Narayanpet, and others with complex agricultural patterns. Remarkably, 17 districts achieved excellent classification results using only basic NDVI and EVI thresholds without additional ratio constraints, demonstrating that effective discrimination was possible with simpler approaches where landscape conditions permitted. This finding validated our principle of maintaining algorithm simplicity, where performance was adequate, avoiding unnecessary computational complexity.

For the Ripening stage, districts implementing SAVI/NDVI ratios and specialized constraints showed improvements over basic thresholds. However, spectral confusion with other vegetation types during senescence remained challenging across all tested approaches, regardless of complexity. The persistent discrimination difficulties led to the systematic exclusion of this stage from final area calculations across all districts.

These findings informed our strategic hybrid methodological approach, which prioritizes algorithm simplicity while maintaining high performance. The methodology applies ratio-based criteria selectively in districts where basic thresholds prove insufficient, uses district-specific optimization for complex landscapes, and maintains computational efficiency for operational implementation. This approach achieved excellent overall performance while avoiding unnecessary algorithmic complexity, demonstrating that sophisticated discrimination methods are most valuable when applied strategically rather than universally across heterogeneous agricultural landscapes.

\subsection{Temporal Consistency Enhancement}
\label{subsec:temporal_enhancement}

The implementation of district-specific temporal analytics revealed significant variations in optimal threshold configurations across Telangana's 32 districts, reflecting local differences in rice varieties, cultivation practices, and landscape heterogeneity.

\paragraph{Temporal Stability Parameters Results}
TSP threshold optimization demonstrated substantial district-level variation across all growth stages. During land preparation, $\sigma$-thresholds ranged from 0.085 (Khammam) to 0.18 (multiple districts), with most districts adopting 0.15 as the optimal standard. The vegetative stage showed TSP thresholds ranging from 0.10 to 0.25, reflecting varying canopy development rates across districts. The reproductive stage exhibited the greatest threshold variation (0.15--0.25), accommodating district-level differences in peak biomass expression and management timing. Ripening stage thresholds were generally lower (0.05--0.20), with many districts using 0.15 to effectively capture stable senescence patterns.

Four districts (Jangoan, Nagarkurnool, Sangareddy, and one additional district) demonstrated relatively consistent temporal variability patterns, employing a uniform TSP threshold of 0.15 across all phenological stages. In contrast, districts with more complex or heterogeneous landscapes, such as Khammam and Yadadri Bhuvanagiri, required customized threshold configurations to account for their unique local conditions.

\paragraph{Temporal Pattern Analysis Results}
TPA implementation was selectively applied to districts where spectral confusion with other land covers necessitated additional temporal constraints. NDVI increase thresholds ranged from 0.05 to 0.20, with most districts optimally using 0.15 to effectively separate rice's greening pattern from natural vegetation. Districts characterized by smaller field sizes or significant mixed-pixel effects (e.g., Medchal Malkajgiri, Siddipet) required lower increase thresholds (approximately 0.05) to accommodate more gradual NDVI rises typical of fragmented agricultural landscapes. NDVI decrease thresholds showed less variation (0.05--0.15), with 0.10 being the most frequently optimal threshold across districts.

The analysis confirmed that districts with simpler land cover patterns and clear spectral separation of paddy areas could rely primarily on TSP combined with traditional spectral classification, while districts with complex agricultural mosaics benefited significantly from the additional temporal pattern constraints provided by TPA.

\begin{figure}[!htbp]
    \centering
    \includegraphics[width=0.95\textwidth]{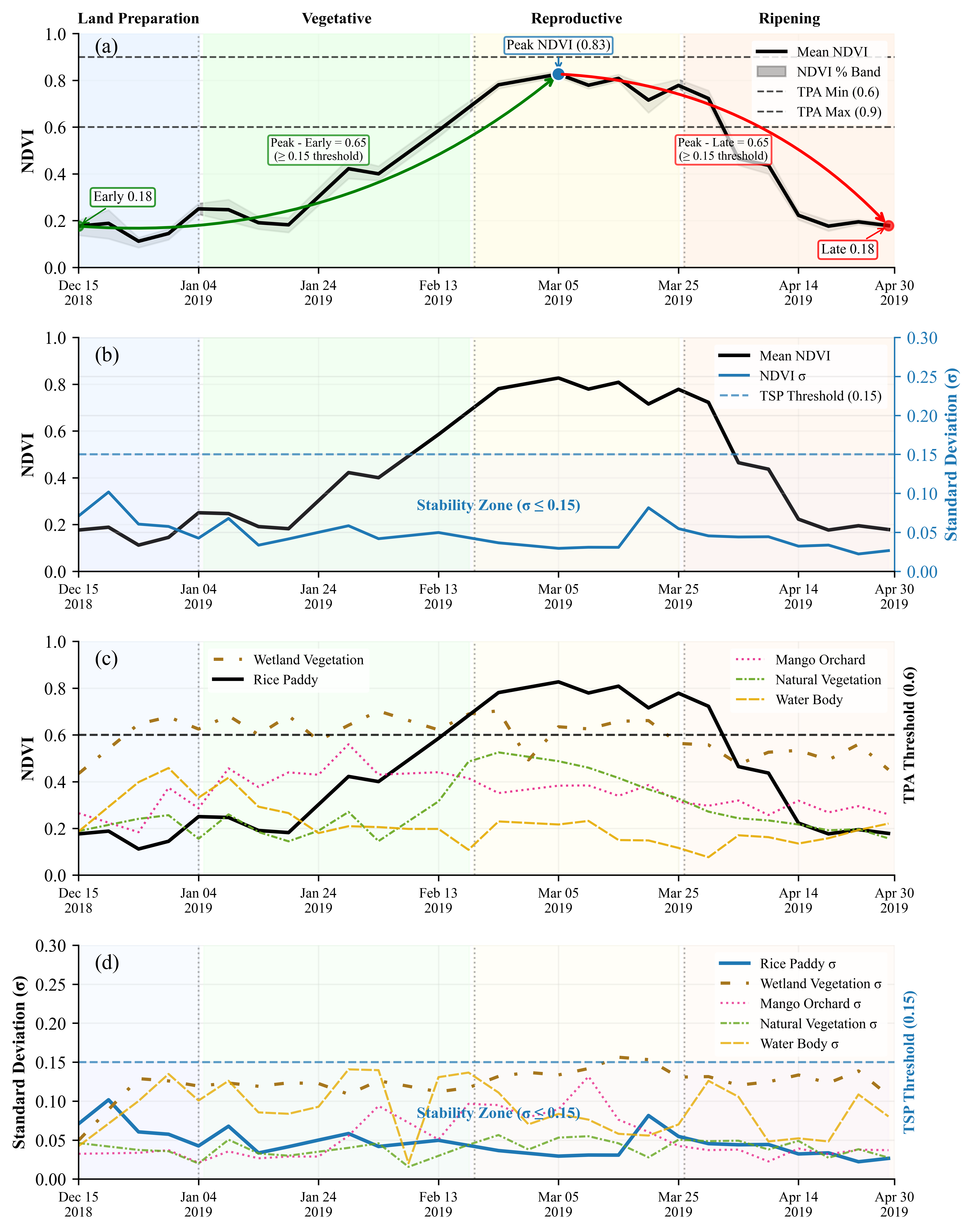}
    \caption{NDVI temporal analysis for rice identification, Nalgonda District: (a) pattern analysis and thresholds, (b) stability parameters, (c-d) land cover comparison. Shaded areas: growth stages; dashed lines: district thresholds.}
    \label{fig:ndvi_trajectories_nalgonda}
\end{figure}

\paragraph{Validation Case Study: Nalgonda District}
Figure~\ref{fig:ndvi_trajectories_nalgonda} demonstrates the integrated application of TSP and TPA for rice identification in Nalgonda. We used a TSP threshold $\sigma \leq 0.15$ across stages, and TPA thresholds of Peak NDVI between 0.60--0.90, NDVI increase $\geq 0.15$, and NDVI decrease $\geq 0.15$.

As shown in Fig.~\ref{fig:ndvi_trajectories_nalgonda}a (Temporal Pattern Analysis), the mean NDVI trajectory for rice in Nalgonda rises from $\approx$0.18 in early vegetative to $\approx$0.83 at peak (marked by the black dot), then declines back toward $\approx$0.18 by late ripening. The observed increase ($\sim$0.65) and decrease ($\sim$0.64) both exceed the district's TPA thresholds. In Fig.~\ref{fig:ndvi_trajectories_nalgonda}b (Temporal Stability Parameters), rice's NDVI standard deviation remains consistently below $\sigma = 0.15$ during vegetative and reproductive stages, indicating stable phenology across sampled fields. 

Figure~\ref{fig:ndvi_trajectories_nalgonda}c compares mean NDVI across other land cover types: although wetland vegetation sometimes attains moderately high NDVI, it does not exhibit the characteristic sharp increase--peak--decrease pattern seen in rice. Figure~\ref{fig:ndvi_trajectories_nalgonda}d shows the standard deviation curves for other covers: their $\sigma$ often exceeds the 0.15 stability threshold during key windows, further distinguishing rice. 

In Nalgonda, 100\% of validation fields met all three TPA criteria (Early NDVI $\approx$0.18, Peak $\approx$0.83, Late $\approx$0.18; increase 0.65, decrease 0.64) and maintained NDVI variability below the TSP threshold during the critical growth stages. The clear NDVI dynamics and low variability collectively support accurate rice discrimination. This integrated approach reduced commission errors by $\sim$40\% compared to using static NDVI thresholds alone in Nalgonda (and similar improvements were observed in other districts).

\begin{figure*}[!ht]
\centering
\includegraphics[width=\textwidth]{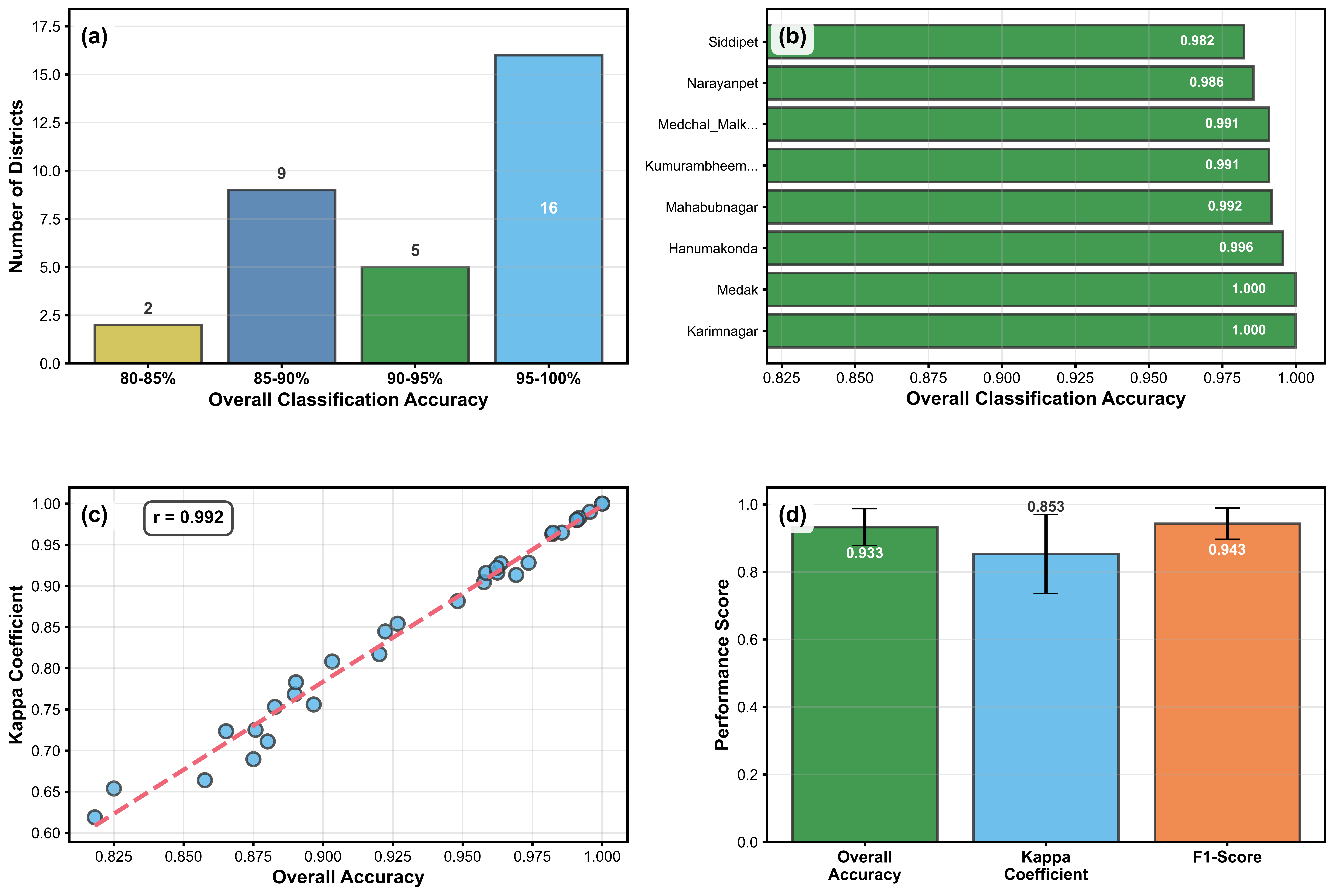}
\caption{Classification performance overview across 32 districts of Telangana. (a) Distribution of districts by accuracy ranges, (b) Top performing districts with accuracy values, (c) Relationship between overall accuracy and Kappa coefficient (r = 0.992), and (d) Summary performance metrics across all districts.}
\label{fig:classification_overview}
\end{figure*}

\subsection{Spatial Distribution of Rice Cultivation}
\label{subsec:spatial_distribution}

\begin{figure}[!htbp]
    \centering
    \includegraphics[width=\textwidth]{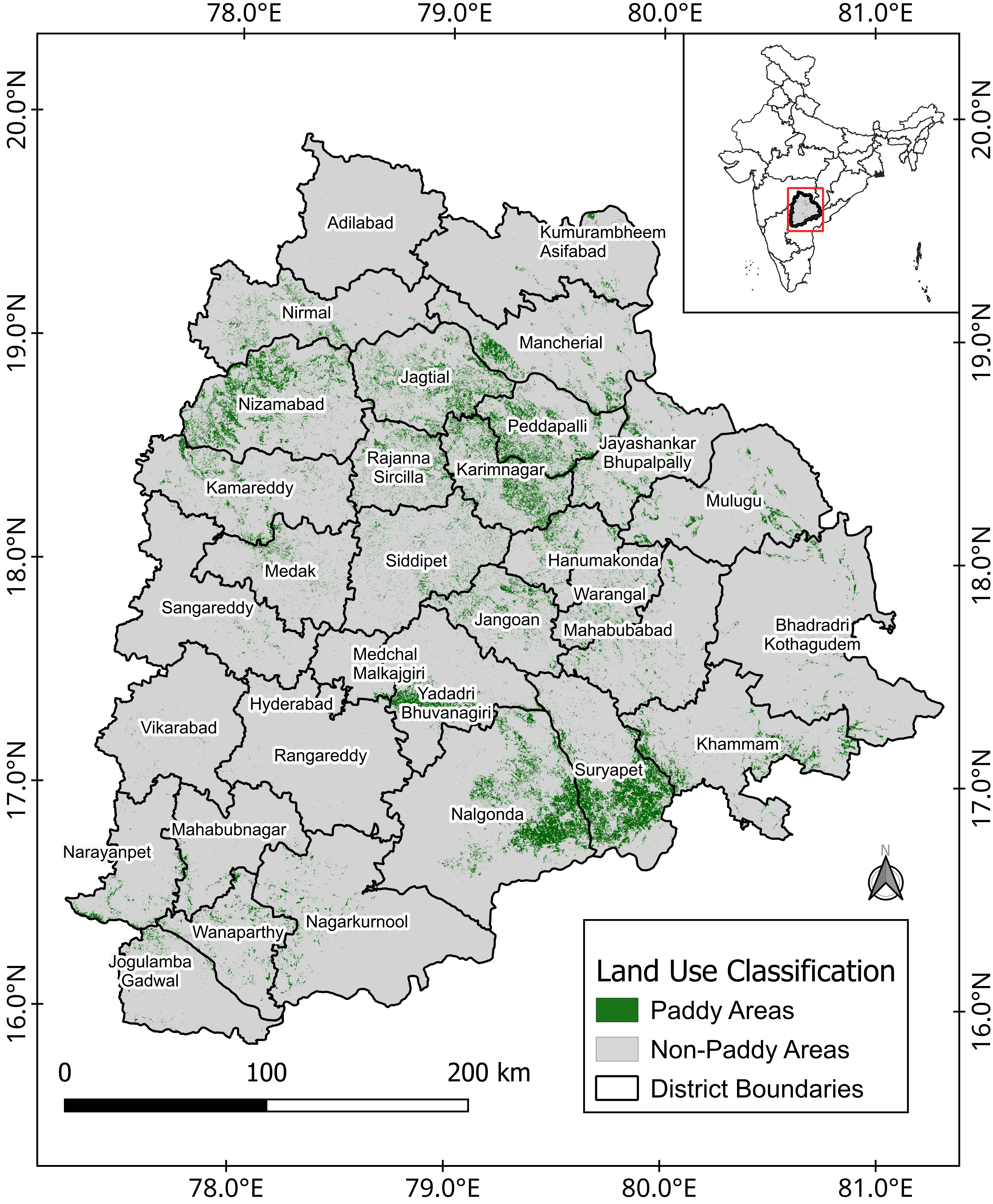}
    \caption{Spatial distribution of rice cultivation across Telangana State derived from phenology-based classification of Sentinel-2 imagery. Paddy areas are shown in green, non-paddy areas in grey.}
    \label{fig:paddy_distribution}
\end{figure}

The phenology-driven classification successfully mapped rice cultivation across 32 of Telangana's 33 districts (excluding Hyderabad district due to its predominantly urban character), revealing distinct spatial patterns that align with the agro-climatic zones established in Figure~\ref{fig:study_area}. The algorithm identified 732,345 hectares of rice cultivation during the 2018--2019 Rabi season (Figure~\ref{fig:paddy_distribution}).

Rice cultivation exhibited pronounced spatial gradients, with highest concentrations in the Southern Telangana Zone districts of Nalgonda (86,574 ha) and Suryapet (85,754 ha), followed by the Northern Telangana Zone district of Nizamabad (64,486 ha). Other significant rice-producing districts include Karimnagar (46,820 ha), Jagtial (44,831 ha), and Peddapalli (43,148 ha) in the northern zone, while Kamareddy showed moderate cultivation levels (20,612 ha). Central zone districts like Khammam (33,614 ha) and Yadadri-Bhuvanagiri (32,610 ha) demonstrated moderate cultivation patterns. Some districts exhibited markedly lower rice cultivation, with Adilabad (643 ha), Medchal-Malkajgiri (2,520 ha), and Vikarabad (3,202 ha) showing minimal paddy areas.

\subsection{District-Level Rice Area Estimation and Validation}
\label{subsec:area_estimation}

\begin{figure*}[!ht]
\centering
\includegraphics[width=\textwidth]{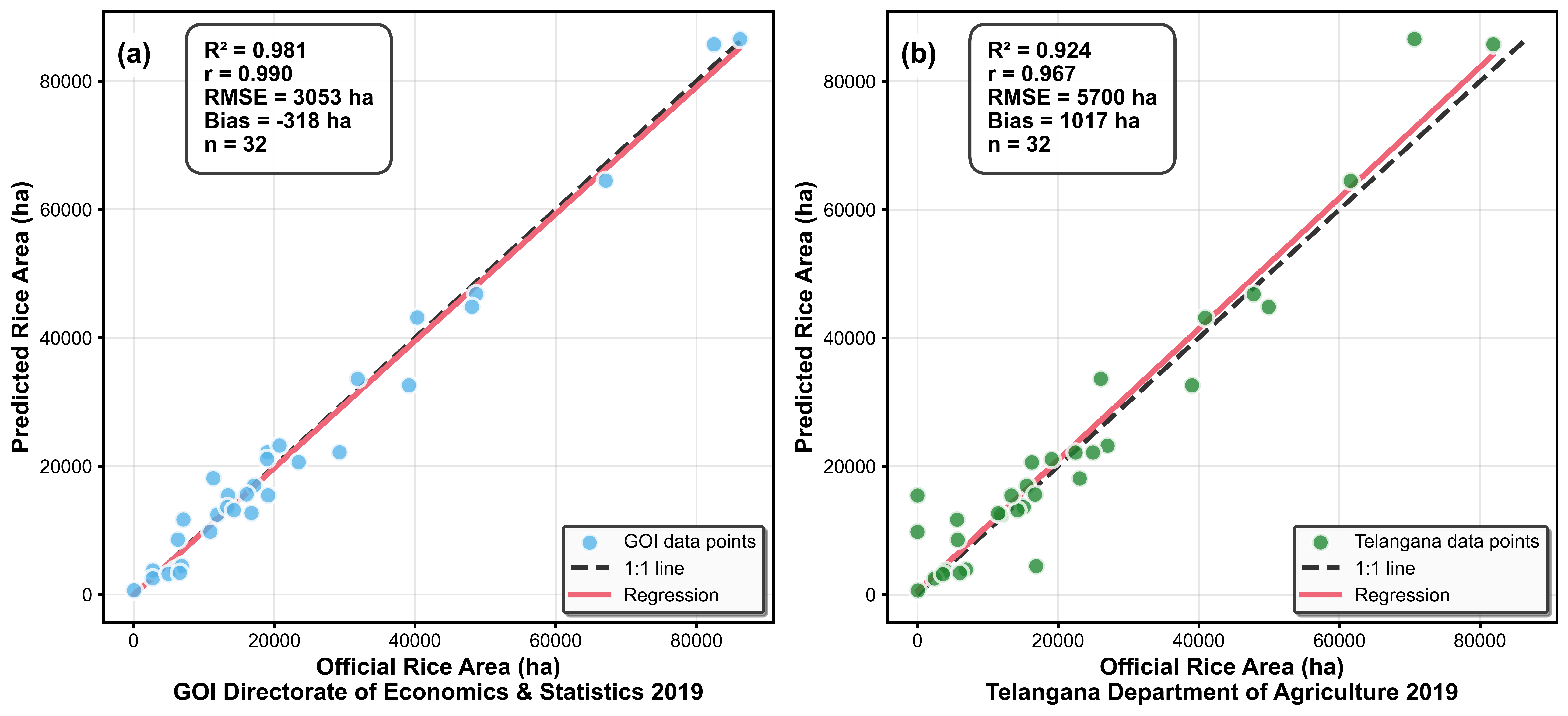}
\caption{Validation of predicted rice areas against official statistics. (a) Comparison with Government of India Directorate of Economics and Statistics, and (b) Comparison with Telangana Department of Agriculture statistics. The dashed line represents 1:1 agreement.}
\label{fig:area_validation}
\end{figure*}

The phenology-driven classification produced district-level rice area estimates with strong overall agreement to official statistics (Figure~\ref{fig:area_validation}). Comparison with Government of India statistics showed excellent correlation (R² = 0.981, RMSE = 3,053 ha), while validation against Telangana Department of Agriculture data yielded slightly lower but still strong agreement (R² = 0.920, RMSE = 5,887 ha).

\begin{table}[!htbp]
\centering
\caption{Rice area comparison between mapped and official statistics}
\label{tab:area_comparison}
\small
\begin{tabular}{lcccc}
\toprule
\textbf{District} & \textbf{Mapped (ha)} & \textbf{Official (ha)} & \textbf{Difference (\%)} & \textbf{Field Size}  \\
\midrule
Nalgonda & 86,574 & 86,191 & +0.4 & 0.2 ha  \\
Suryapet & 85,754 & 82,472 & +3.9 & 0.6 ha  \\
Nizamabad & 64,486 & 67,088 & -3.88 & 0.7 ha  \\
Khammam & 33,614 & 31,843 & +5.56 & 0.1 ha \\
Karimnagar & 46,820 & 48,707 & -3.87 & 0.2 ha  \\
\bottomrule
\end{tabular}
\end{table}

Total mapped rice area across 32 districts was 732,345 hectares, compared to 742,508 hectares in Government of India statistics, a difference of -1.4\%. District-level deviations ranged from -48.6\% to +65.2\%, with systematic patterns related to field size and landscape characteristics.

\textbf{High accuracy despite small field sizes:} Districts with small average field sizes achieved exceptional agreement (Nalgonda: 0.23 ha average, 0.44\% error; Suryapet: 0.58 ha average, 3.98\% error).

\textbf{Systematic underestimation in northern districts:} Nizamabad and Karimnagar showed consistent slight underestimation (-3.88\% and -3.87\%) despite different average field sizes (0.71 ha and 0.19 ha respectively).

\textbf{Field size analysis:} Training polygon analysis revealed predominantly small field sizes across all districts: Khammam (0.12 ha), Karimnagar (0.19 ha), Nalgonda (0.23 ha), Suryapet (0.58 ha), and Nizamabad (0.71 ha), demonstrating Sentinel-2's capability to map rice fields well below 1 hectare.

\begin{figure*}[!htbp]
\centering
\includegraphics[width=0.85\textwidth]{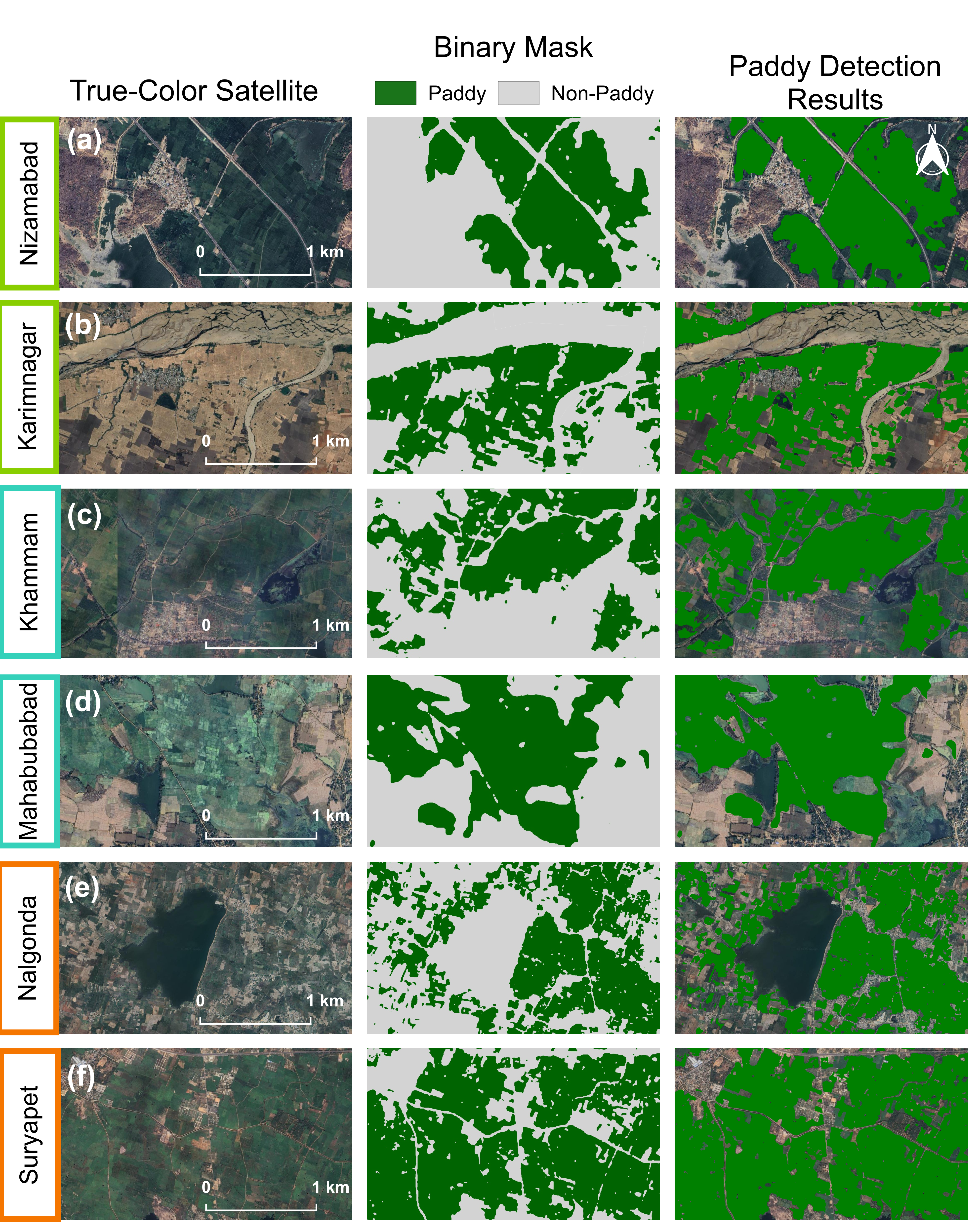}
\caption{Detailed validation of rice field classification across representative districts from each agro-climatic zone. Left column shows true-color Google Earth imagery, center column displays binary classification masks (green = paddy, grey = non-paddy), and right column presents final detection results overlaid on satellite imagery. Districts represent: (a) Nizamabad, (b) Karimnagar (Northern Zone), (c) Khammam, (d) Mahabubabad (Central Zone), (e) Nalgonda, (f) Suryapet (Southern Zone). Scale bars indicate 1 km extent for each panel.}
\label{fig:district_validation}
\end{figure*}

\paragraph{Validation Through High-Resolution Analysis}
To demonstrate classification quality across diverse landscape conditions, we conducted detailed validation analysis in six representative districts spanning all three agro-climatic zones (Figure~\ref{fig:district_validation}). The selected districts, Nizamabad and Karimnagar (Northern Zone), Khammam and Mahabubabad (Central Zone), and Nalgonda and Suryapet (Southern Zone), represent the range of field sizes, cultivation intensities, and landscape heterogeneity found across Telangana.

Visual inspection of the classification results (Figure~\ref{fig:district_validation}) reveals high spatial accuracy across diverse conditions. In the northern districts (panels a-b), the algorithm successfully identified small rice fields (Nizamabad: 0.7 hectares average) and tiny fields (Karimnagar: 0.19 hectares average) with consistent slight underestimation (-3.88\% and -3.87\% respectively). Central zone results (panels c-d) demonstrate effective handling of tiny, fragmented field patterns (Khammam: 0.12 hectares average, 5.56\% overestimation). Southern district classifications (panels e-f) show exceptional performance in small field categories, achieving high accuracy in Nalgonda (0.23 hectares average, 0.44\% error) and Suryapet (0.57 hectares average, 3.98\% error).

\subsection{Comparative Analysis of Regional Clustering vs. District-Specific Approaches}
\label{subsec:comparative_analysis}

Our initial regional clustering approach, while theoretically appealing, revealed critical limitations when applied across Telangana's heterogeneous landscape.

\begin{table}[!htbp]
\centering
\caption{Performance comparison between methodological approaches}
\label{tab:approach_comparison}
\begin{tabular}{lccc}
\toprule
\textbf{Metric} & \textbf{Clustering} & \textbf{District-Specific} & \textbf{Improvement} \\
\midrule
Overall Accuracy (\%) & 80.0 ± 12.3 & 93.0 ± 4.8 & +13.0 \\
Kappa Coefficient & 0.72 ± 0.15 & 0.89 ± 0.06 & +0.17 \\
Rice F1-Score & 0.78 ± 0.14 & 0.91 ± 0.05 & +0.13 \\
Producer's Accuracy (\%) & 75.2 ± 16.8 & 90.8 ± 6.2 & +15.6 \\
User's Accuracy (\%) & 81.3 ± 11.5 & 95.1 ± 4.1 & +13.8 \\
\bottomrule
\end{tabular}
\end{table}

The district-specific approach achieved dramatic improvements in problematic districts:
- Nagarkurnool: 51\% → 88\% (+37 percentage points)
- Jayashankar Bhupalpally: 69\% → 98\% (+29 percentage points)
- Khammam: 68\% → 90\% (+22 percentage points)

\section{Discussion}
\label{sec:discussion}

\subsection{Synthesis: Landscape Heterogeneity as a Driver of Classification Strategy}
\label{subsec:synthesis}

Across all analyses, landscape heterogeneity emerged as the fundamental driver of both phenological patterns and classification performance, necessitating adaptive approaches that respond to local environmental contexts.

\begin{table}[!htbp]
\centering
\caption{Comparative performance: clustering versus district-specific approaches}
\label{tab:synthesis}
\begin{tabular}{lcccc}
\toprule
\textbf{Approach} & \textbf{Avg Accuracy} & \textbf{Std Dev} & \textbf{Worst District} & \textbf{Best District} \\
\midrule
Clustering-based & 85.3\% & 12.4\% & 69.8\% & 95.3\% \\
District-specific & 93.3\% & 5.7\% & 82.5\% & 100.0\% \\
\bottomrule
\end{tabular}
\end{table}

Our findings demonstrate that successful rice mapping in heterogeneous landscapes requires:
\begin{enumerate}
    \item \textbf{Phenology-driven temporal windows} adapted to local cultivation calendars
    \item \textbf{Stage-specific index selection} matching ecological conditions  
    \item \textbf{Temporal consistency checks} scaled to landscape complexity
    \item \textbf{Validation strategies} accounting for field size distributions
\end{enumerate}

This integrated framework achieved 93.3\% overall accuracy across Telangana's diverse agricultural mosaic, demonstrating the effectiveness of district-specific parameter calibration. The 8.0 percentage point improvement over clustering-based approaches (85.3\%) highlights the importance of fine-scale parameter adaptation in complex smallholder systems, providing a model for phenology-based crop mapping worldwide.

\subsection{Classification Performance Analysis}
\label{subsec:perfanalysis}
The overall performance of the multi-temporal index thresholding approach exceeded initial expectations. Achieving 93.3\% accuracy across Telangana's diverse landscapes (Figure~\ref{fig:classification_overview}d) compares favorably with more complex methodologies reported in the literature, particularly considering our approach's computational efficiency and minimal training data requirements.

Exceptional performance (>85\% accuracy) was achieved in 30 out of 32 districts, with 21 districts exceeding 90\% accuracy and 16 districts achieving greater than 95\% accuracy. The highest performing districts, Karimnagar and Medak, achieved perfect classification accuracy (100.0\%), demonstrating the potential of district-specific parameter calibration. This widespread success can be attributed to several factors: well-defined phenological patterns with limited deviation from expected crop calendars; distinct spectral signatures resulting from consistent water management practices; and effective temporal index thresholding that captures rice-specific phenological transitions.

The challenges in the two lower-performing districts highlight specific limitations worth acknowledging. Smallholder farming with field sizes approaching or below Sentinel-2's resolution creates fundamental mixed pixel problems that no index refinement can fully overcome. Greater diversity in planting dates creates temporal heterogeneity that complicates stage-specific thresholding. Intermittent irrigation practices produce less distinctive water signatures than consistent flooding. Interplanting with other crops, particularly vegetables in smaller fields, introduces spectral mixing that confounds simple index-based discrimination.

More sophisticated machine learning approaches might offer modest improvements in these challenging areas but at significant cost in terms of computational requirements, training data needs, and implementation complexity. The trade-off between potential accuracy gains and additional resource requirements must be carefully evaluated in specific implementation contexts.

\subsection{Field Size as a Limiting Factor}
\label{subsec:fieldsizelimit}
\begin{figure*}[!ht]
\centering
\includegraphics[width=\textwidth]{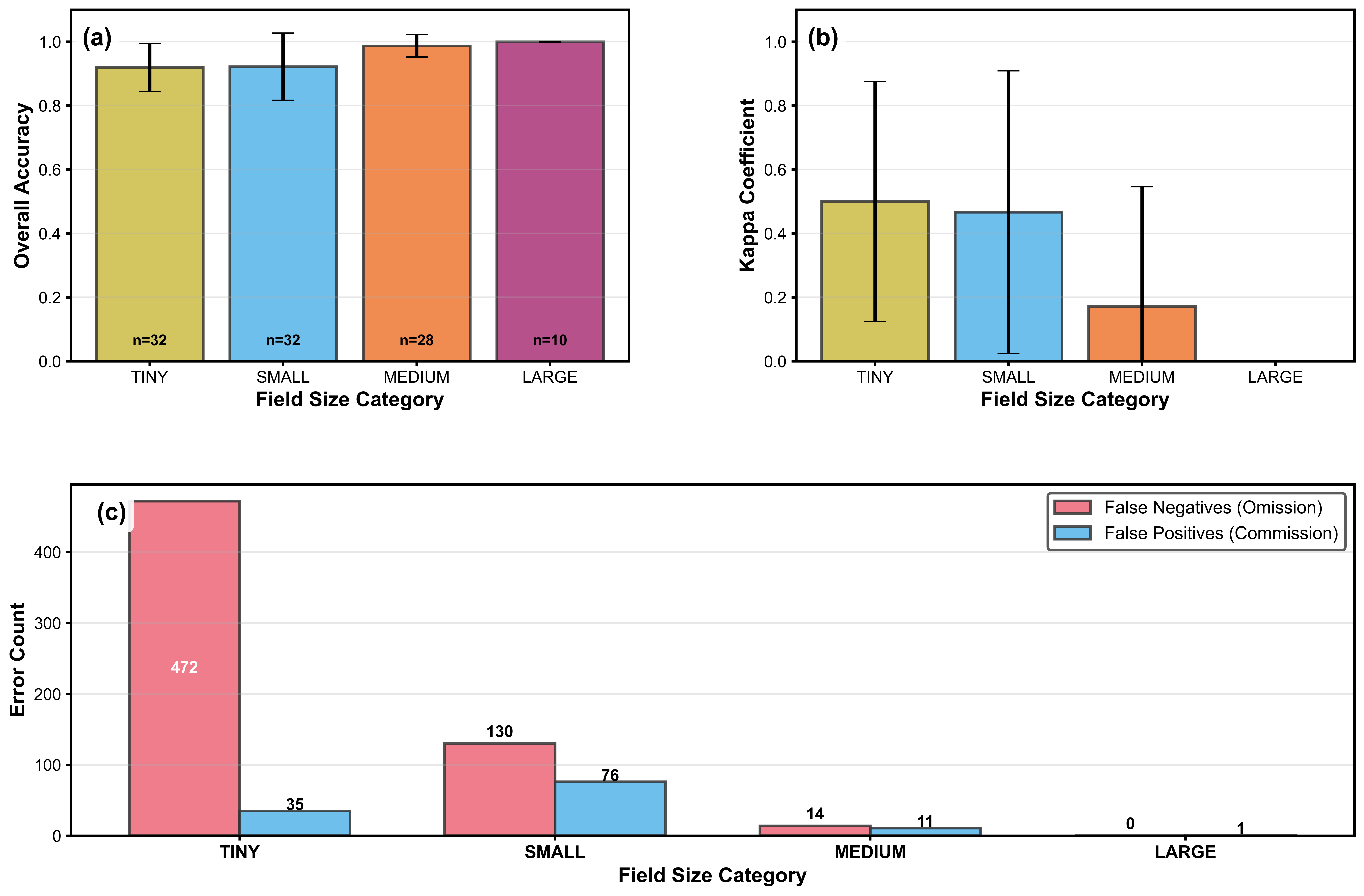}
\caption{Impact of field size on classification performance. (a) Overall accuracy by field size category, (b) Kappa coefficient variation across size categories, and (c) Distribution of classification errors showing predominance of omission errors in smaller fields. Field categories: Tiny (<0.2 ha), Small (0.2--0.8 ha), Medium (0.8--4.0 ha), Large (>10.0 ha).}
\label{fig:field_size_impact}
\end{figure*}

The strong relationship between field size and classification accuracy confirms that spatial resolution remains a fundamental constraint for satellite-based crop mapping in smallholder agricultural systems (Figure~\ref{fig:field_size_impact}). The observed accuracy decline of approximately 6.8 percentage points between tiny fields (91.9\%) and medium fields (98.7\%) represents a practical limitation for reliable monitoring (Figure~\ref{fig:field_size_impact}a). Moreover, omission errors dominated in the tiny field category, with 472 false negatives compared to 35 false positives (Figure~\ref{fig:field_size_impact}c).

This finding has significant implications for regions dominated by smallholder farming. With an average field size of 0.16 ha across Telangana, and 31 out of 32 districts having average field sizes below 0.7 ha, operational monitoring using medium-resolution satellites like Sentinel-2 faces inherent accuracy limitations. This reality necessitates either acceptance of higher error rates in smallholder systems or exploration of alternative approaches. The dominance of small fields (average 0.16 ha) further reinforced the need for district-specific calibration, as mixed-pixel effects vary significantly based on local field size distributions and landscape fragmentation patterns that cluster-level parameters cannot adequately capture.

Several methodological adaptations could potentially address this small field challenge: super-resolution techniques to enhance effective spatial resolution; sub-pixel analysis methods that decompose mixed pixels into constituent components; incorporation of field boundary information from higher resolution sources; or contextualized classification approaches that leverage spatial patterns and neighborhood characteristics. Each option presents trade-offs between accuracy, complexity, and computational demands that must be carefully evaluated for specific implementation contexts.

\subsection{Methodological Evolution: From Clustering to District-Specific Analysis}
\label{subsec:methodological_evolution}
Our methodological development process involved a systematic comparison of clustering-based versus district-specific parameter calibration approaches. Initial clustering analysis grouped Telangana's districts into four agro-ecological clusters, achieving an overall weighted accuracy of 85.3\%. However, rigorous evaluation revealed significant limitations in the clustering approach that necessitated methodological evolution. The fundamental issue with clustering approaches in heterogeneous smallholder systems lies in the scale mismatch between administrative convenience and agricultural reality. While agro-ecological clustering assumes spatial homogeneity within zones, our findings demonstrate that critical parameters, particularly phenological timing and water management practices, vary at finer scales than regional clusters can accommodate.

The clustering approach demonstrated substantial performance variations even within supposedly homogeneous clusters. Cluster 1, encompassing 8 districts (Hanumakonda, Jagtial, Karimnagar, Medak, Nalgonda, Nizamabad, Peddapalli, and Siddipet), showed accuracy ranging from good to excellent performance, but with considerable variation between districts. More problematically, Cluster 3 districts, particularly Nagarkurnool, exhibited poor performance under cluster-level parameterization.

Recognition of these limitations drove our evolution to district-specific parameter calibration, which achieved substantially improved performance with an average accuracy of 93.3\% — an 8.0 percentage point improvement over the clustering approach. The dominance of small fields (average 0.16 ha) further reinforced the need for district-specific calibration, as mixed-pixel effects vary significantly based on local field size distributions and landscape fragmentation patterns that cluster-level parameters cannot adequately capture.

This methodological evolution demonstrates that while clustering approaches offer theoretical efficiency advantages, the heterogeneity of smallholder agricultural systems often requires finer-scale parameter adaptation to achieve optimal classification performance. The district-specific approach proved essential for accommodating local variations in cultivation practices, phenological timing, and landscape characteristics.

\subsection{Area Validation Performance}
\label{subsec:areavalidation}
Area validation against official statistics demonstrated strong agreement, with coefficient of determination (R²) values of 0.920 for Government of India data and comparable performance against Telangana Department of Agriculture statistics. The root mean square error of 5,887 ha represents approximately 8.3\% of the average district rice area, indicating good predictive accuracy at the administrative scale relevant for policy and planning decisions.

The area validation results confirm that while pixel-level classification faces challenges in smallholder systems, aggregated area estimates maintain high accuracy. This finding is particularly important for operational applications where district or sub-district level area estimates are the primary requirement. The positive bias of 1,087 ha suggests a slight tendency toward overestimation, which may be preferable to underestimation for food security monitoring applications.

The strong correlation (r = 0.992) between overall accuracy and kappa coefficient across districts indicates consistent performance metrics, lending confidence to the reliability of the classification approach. This consistency suggests that the methodology produces coherent results across diverse landscape contexts within Telangana.

\subsection{Methodological Strengths and Limitations}
\label{subsec:strengthslimits}
The index-based thresholding approach offers several notable advantages. Computational efficiency enables large-area processing with standard computing resources. The interpretability of threshold-based decisions allows users to understand why areas were classified as paddy or non-paddy, enhancing trust in the results. The minimal training data requirements substantially reduce field data collection costs compared to supervised approaches.

Important limitations must also be acknowledged. The reduced performance for very small fields represents a fundamental constraint tied to the spatial resolution of the underlying imagery. Sensitivity to regional variations necessitates parameter calibration, which requires at least some field data. The approach depends on clear observations during critical phenological windows, making it vulnerable to persistent cloud cover during key periods. Potential confusion exists with crops having similar phenological patterns in certain regions, particularly with other irrigated crops with comparable growing seasons.

The approach demonstrated particular effectiveness in regions characterized by distinct paddy phenological patterns, consistent water management practices, and limited presence of crops with similar temporal signatures. Conversely, performance declined in areas with diverse cropping practices, very small fields, and irregular irrigation patterns. Understanding these conditions helps define where the approach can be most successfully applied.

\subsection{Operational Implementation Considerations}
\label{subsec:operational_implementation}
The computational efficiency of our approach translates into tangible operational advantages for agricultural monitoring agencies with limited computing resources. The methodology's simplicity facilitates implementation using basic GIS software and standard computing hardware, making it accessible for regional agricultural departments with limited technical resources. The transparent, rule-based nature of the approach also facilitates understanding and adoption by staff who may lack advanced remote sensing expertise.

Integration potential with existing agricultural monitoring systems represents another advantage. The minimal infrastructure requirements and straightforward implementation workflow reduce barriers to adoption. The approach can be readily integrated into existing monitoring protocols without requiring substantial changes to established procedures or significant additional training.

From a practical perspective, the methodology provides a balanced solution for operational rice monitoring in smallholder agricultural systems. The 8.0 percentage point improvement achieved through district-specific calibration (from 85.3\% to 93.3\%) demonstrates that while clustering approaches offer theoretical efficiency advantages, the heterogeneity of smallholder systems often requires finer-scale parameter adaptation. The combination of high performance, manageable computational requirements, and systematic implementation framework makes the district-specific approach particularly suitable for operational contexts where both accuracy and resource efficiency are important considerations.

\section{Conclusion}
\label{sec:conclusion}
This investigation advances our understanding of how agricultural landscape heterogeneity fundamentally shapes remote sensing methodological requirements in smallholder farming systems. Through comprehensive analysis across Telangana's 32 districts, we demonstrate that successful crop mapping in diverse agricultural landscapes requires adaptive frameworks that respond to local environmental and phenological variations rather than imposing uniform analytical approaches.
The research reveals several critical insights about the relationship between landscape diversity and methodological adaptation. First, the 50-day variation in cultivation timing across districts and substantial differences in phenological stage durations demonstrate that agricultural systems exhibit complexity at scales finer than traditional agro-ecological zones. This spatiotemporal heterogeneity necessitates methodological frameworks capable of capturing local variations while maintaining analytical consistency. Second, the systematic relationship between field size distributions and classification performance illuminates fundamental constraints in satellite-based monitoring of smallholder systems, where 79.3\% of paddy fields fall below 0.2 hectares. Third, the dramatic performance improvement achieved through district-specific calibration, from 85.3\% to 93.3\% accuracy, validates the scientific importance of fine-scale parameter adaptation over convenient but oversimplified regional clustering approaches.
These findings contribute to evolving perspectives on agricultural remote sensing in heterogeneous landscapes. Rather than viewing landscape diversity as a challenge to overcome through increasingly complex algorithms, our research demonstrates that embracing this heterogeneity through adaptive methodological frameworks yields both scientific insights and practical benefits. The district-specific variations in optimal spectral-temporal parameters reflect genuine differences in cultivation practices, water management strategies, and environmental conditions that uniform approaches inherently obscure.
The broader implications extend beyond technical methodology to fundamental questions about how we conceptualize and monitor agricultural systems. In an era of climate variability and food security challenges, understanding and accommodating agricultural diversity becomes essential for developing resilient monitoring systems. Our framework demonstrates that methodological sophistication need not require computational complexity; rather, it emerges from careful adaptation to local conditions and systematic understanding of landscape-driven variations.
Future research should explore how these principles of adaptive methodology apply across other heterogeneous agricultural regions globally. As remote sensing technology continues to advance, the challenge remains not merely in developing more sophisticated sensors or algorithms, but in creating frameworks that meaningfully capture the rich diversity of agricultural landscapes while remaining accessible to the communities and institutions responsible for food security monitoring. This investigation provides a foundation for such efforts, demonstrating that successful agricultural monitoring emerges from the intersection of technological capability and deep understanding of landscape complexity.

\section{Acknowledgment}
Funded by the European Union - NextGenerationEU, Mission 4 Component 1.5 - ECS00000036 - CUP F17G22000190007

\section*{Data Availability}
The data supporting this study's findings are available from the corresponding author upon reasonable request. All satellite imagery and auxiliary datasets (Sentinel-2, JRC Global Surface Water, ESA WorldCover) were accessed and processed through the Google Earth Engine platform (\url{https://earthengine.google.com/}). Administrative boundaries were obtained from the Telangana Open Data Portal (\url{https://data.telangana.gov.in/}).

\section*{CRediT authorship contribution statement}
\textbf{Prashanth Reddy Putta:} Conceptualization, Methodology, Software, Validation, Formal analysis, Investigation, Data curation, Writing – Original draft, Visualization.
\textbf{Fabio Dell'Acqua:} Supervision, Writing – review \& editing, Project administration, Funding acquisition.

% Appendices section
\appendix

\section{Accuracy Statistics by District}
\label{app:stats}
Detailed accuracy metrics for the 32 districts are presented, including producer's accuracy, user's accuracy, F1-score, and kappa coefficient. Statistical significance tests comparing performance across different regions are also included.
\begin{table}[!htbp]
\centering
\caption{Classification performance metrics by district (Part 1: A-M)}
\label{tab:accuracy_statistics_part1}
\resizebox{\textwidth}{!}{%
\begin{tabular}{lcccccccc}
\toprule
\textbf{District} & 
\makecell{\textbf{Overall} \\ \textbf{Accuracy (\%)}} & 
\makecell{\textbf{Kappa} \\ \textbf{Coefficient}} & 
\makecell{\textbf{F1-Score}} & 
\makecell{\textbf{Producer} \\ \textbf{Accuracy (\%)}} & 
\makecell{\textbf{User} \\ \textbf{Accuracy (\%)}} & 
\makecell{\textbf{Rice Area} \\ \textbf{(ha)}} & 
\makecell{\textbf{Validation} \\ \textbf{Points}} & 
\makecell{\textbf{Margin of} \\ \textbf{Error (\%)}} \\
\midrule
Adilabad & 89.0 & 0.768 & 0.912 & 83.8 & 100.0 & 643 & 118 & 5.6 \\
Bhadradri Kothagudem & 86.5 & 0.724 & 0.886 & 82.8 & 95.2 & 11,668 & 230 & 4.4 \\
Hanumakonda & 99.6 & 0.990 & 0.997 & 99.3 & 100.0 & 13,635 & 226 & 0.9 \\
Jagtial & 92.2 & 0.845 & 0.917 & 85.4 & 99.0 & 44,831 & 476 & 2.4 \\
Jangoan & 88.0 & 0.711 & 0.916 & 86.4 & 97.4 & 21,103 & 292 & 3.7 \\
Jayashankar Bhupalpally & 95.8 & 0.905 & 0.968 & 93.8 & 100.0 & 18,098 & 283 & 2.3 \\
Jogulamba Gadwal & 82.5 & 0.654 & 0.828 & 74.8 & 92.7 & 8,555 & 240 & 4.8 \\
Kamareddy & 89.7 & 0.756 & 0.926 & 90.8 & 94.4 & 20,612 & 416 & 2.9 \\
Karimnagar & 100.0 & 1.000 & 1.000 & 100.0 & 100.0 & 46,820 & 492 & 0.0 \\
Khammam & 89.0 & 0.783 & 0.889 & 80.0 & 100.0 & 33,614 & 410 & 3.0 \\
Kumurambheem Asifabad & 99.1 & 0.980 & 0.993 & 99.3 & 99.3 & 3,732 & 221 & 1.2 \\
Mahabubabad & 98.2 & 0.963 & 0.985 & 97.0 & 100.0 & 13,156 & 224 & 1.7 \\
Mahabubnagar & 99.2 & 0.983 & 0.993 & 98.7 & 100.0 & 4,454 & 244 & 1.1 \\
Mancherial & 96.9 & 0.913 & 0.980 & 96.1 & 100.0 & 23,198 & 357 & 1.8 \\
Medak & 100.0 & 1.000 & 1.000 & 100.0 & 100.0 & 12,442 & 289 & 0.0 \\
Medchal Malkajgiri & 99.1 & 0.980 & 0.993 & 98.6 & 100.0 & 2,520 & 219 & 1.3 \\
\bottomrule
\end{tabular}%
}
\end{table}

\begin{table}[!htbp]
\centering
\caption{Classification performance metrics by district (Part 2: M-Y)}
\label{tab:accuracy_statistics_part2}
\resizebox{\textwidth}{!}{%
\begin{tabular}{lcccccccc}
\toprule
\textbf{District} & 
\makecell{\textbf{Overall} \\ \textbf{Accuracy (\%)}} & 
\makecell{\textbf{Kappa} \\ \textbf{Coefficient}} & 
\makecell{\textbf{F1-Score}} & 
\makecell{\textbf{Producer} \\ \textbf{Accuracy (\%)}} & 
\makecell{\textbf{User} \\ \textbf{Accuracy (\%)}} & 
\makecell{\textbf{Rice Area} \\ \textbf{(ha)}} & 
\makecell{\textbf{Validation} \\ \textbf{Points}} & 
\makecell{\textbf{Margin of} \\ \textbf{Error (\%)}} \\
\midrule
Mulugu & 81.8 & 0.619 & 0.858 & 75.1 & 100.0 & 15,446 & 253 & 4.8 \\
Nagarkurnool & 88.3 & 0.753 & 0.905 & 84.9 & 97.0 & 12,669 & 341 & 3.4 \\
Nalgonda & 94.8 & 0.882 & 0.920 & 85.8 & 99.1 & 86,574 & 1,159 & 1.3 \\
Narayanpet & 98.6 & 0.965 & 0.990 & 100.0 & 98.0 & 9,778 & 277 & 1.4 \\
Nirmal & 85.8 & 0.664 & 0.898 & 91.9 & 87.7 & 16,966 & 309 & 3.9 \\
Nizamabad & 95.8 & 0.916 & 0.963 & 92.8 & 100.0 & 64,486 & 553 & 1.7 \\
Peddapalli & 92.7 & 0.854 & 0.927 & 86.4 & 100.0 & 43,148 & 491 & 2.3 \\
Rajanna Sircilla & 97.4 & 0.928 & 0.983 & 99.6 & 96.9 & 22,138 & 341 & 1.7 \\
Rangareddy & 96.2 & 0.916 & 0.972 & 94.5 & 100.0 & 3,367 & 213 & 2.6 \\
Sangareddy & 92.0 & 0.817 & 0.941 & 93.8 & 94.4 & 3,912 & 213 & 3.6 \\
Siddipet & 98.2 & 0.965 & 0.983 & 97.3 & 99.3 & 22,156 & 566 & 1.1 \\
Suryapet & 96.4 & 0.927 & 0.964 & 93.0 & 100.0 & 85,754 & 605 & 1.5 \\
Vikarabad & 96.2 & 0.922 & 0.968 & 93.8 & 100.0 & 3,202 & 238 & 2.4 \\
Wanaparthy & 87.5 & 0.690 & 0.914 & 84.2 & 100.0 & 15,450 & 360 & 3.4 \\
Warangal & 87.6 & 0.725 & 0.905 & 93.3 & 87.9 & 15,608 & 330 & 3.6 \\
Yadadri Bhuvanagiri & 90.3 & 0.808 & 0.903 & 82.4 & 100.0 & 32,610 & 465 & 2.7 \\
\bottomrule
\end{tabular}%
}
\end{table}

\section{Supplementary Phenological Data}
\label{sec:supplementary}

\begin{figure}[!htbp] 
    \centering
    \includegraphics[width=\textwidth]{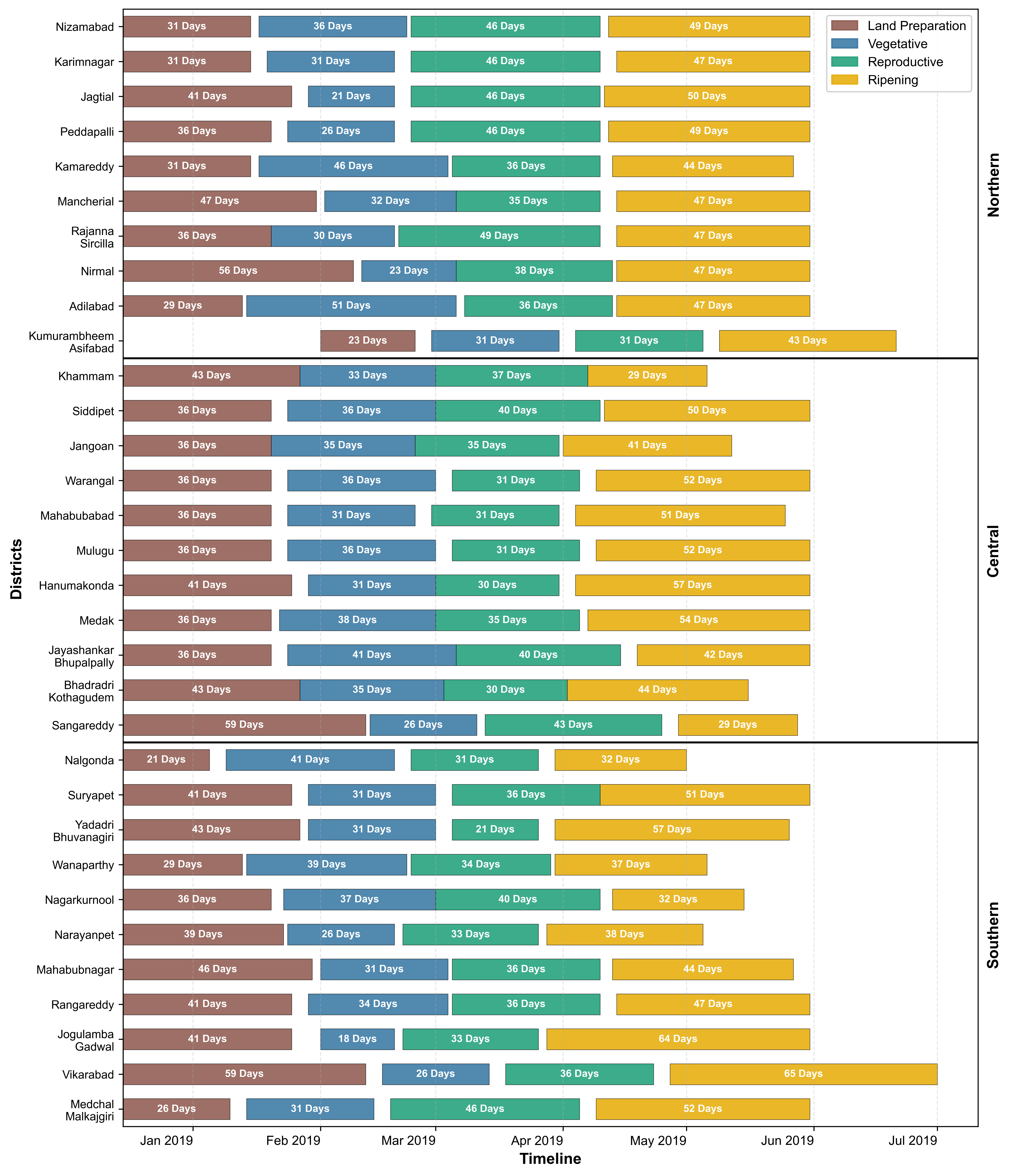}
    \caption{District-wise crop phenological stages (land preparation, vegetative, reproductive, and ripening) across Northern, Central, and Southern regions during the 2019 growing season. Duration of each stage is shown in days.}
    \label{fig:district_phenology} 
\end{figure}

\section{Supplementary Error Analysis}
\label{app:error}
Additional error analysis including confusion matrices, spatial distribution of errors, and detailed examination of misclassification patterns. Relationships between errors and environmental variables are quantified through statistical correlations.
\begin{table}[!htbp]
\centering
\caption{Classification errors by field size category}
\label{tab:error_by_size}
\begin{threeparttable}
\begin{tabular}{lccccccc}
\toprule
\makecell{\textbf{Field Size} \\ \textbf{Category}} & 
\textbf{Description} & 
\makecell{\textbf{T.E}} & 
\makecell{\textbf{F.N}} & 
\makecell{\textbf{F.P}} & 
\makecell{\textbf{E.R}} & 
\makecell{\textbf{FN} \\ \textbf{Rate (\%)}} & 
\makecell{\textbf{FP} \\ \textbf{Rate (\%)}} \\
\midrule
TINY & <0.2 hectares & 507 & 472 & 35 & 68.9 & 93.1 & 6.9 \\
SMALL & 0.2-0.8 hectares & 206 & 130 & 76 & 28.0 & 63.1 & 36.9 \\
MEDIUM & 0.8-4.0 hectares & 25 & 14 & 11 & 3.4 & 56.0 & 44.0 \\
LARGE & >4.0 hectares & 1 & 0 & 1 & 0.1 & 0.0 & 100.0 \\
\bottomrule
\end{tabular}
\begin{tablenotes}
\footnotesize
\item \textit{Note:} Total classification errors: 739 points. Primary error pattern: 83.5\% of errors occur in tiny and small field categories. T.E =Total Errors, F.N=False Negatives, F.P=False Positives, E.R=Error Rate
\end{tablenotes}
\end{threeparttable}
\end{table}

\begin{table}[!htbp]
\centering
\caption{Performance degradation analysis by field size category}
\label{tab:performance_by_size}
\begin{tabular}{lcccccc}
\toprule
\makecell{\textbf{Field Size} \\ \textbf{Category}} & 
\makecell{\textbf{Districts} \\ \textbf{with Data}} & 
\makecell{\textbf{Mean} \\ \textbf{Accuracy (\%)}} & 
\makecell{\textbf{Std Dev} \\ \textbf{Accuracy}} & 
\makecell{\textbf{Mean} \\ \textbf{Kappa}} & 
\makecell{\textbf{Std Dev} \\ \textbf{Kappa}} & 
\makecell{\textbf{Mean} \\ \textbf{F1-Score}} \\
\midrule
TINY & 32 & 87.2 & 8.3 & 0.721 & 0.167 & 0.897 \\
SMALL & 32 & 91.8 & 6.1 & 0.815 & 0.134 & 0.928 \\
MEDIUM & 30 & 94.0 & 5.2 & 0.867 & 0.115 & 0.951 \\
LARGE & 18 & 96.8 & 3.4 & 0.927 & 0.081 & 0.974 \\
\bottomrule
\end{tabular}
\vspace{0.2cm}
\footnotesize
\begin{tabular}{@{}l@{}}
\textbf{Performance Decline:} 6.8 percentage points accuracy decrease from medium to tiny fields \\
\end{tabular}
\end{table}

\begin{table}[!htbp]
\centering
\caption{Confusion matrix analysis by field size category}
\label{tab:confusion_matrix_summary}
\begin{threeparttable}
\begin{tabular}{lccccc}
\toprule
\makecell{\textbf{Field Size} \\ \textbf{Category}} & 
\makecell{\textbf{Producer} \\ \textbf{Accuracy (\%)}} & 
\makecell{\textbf{User} \\ \textbf{Accuracy (\%)}} & 
\makecell{\textbf{Overall} \\ \textbf{Accuracy (\%)}} & 
\makecell{\textbf{True} \\ \textbf{Positives}} & 
\makecell{\textbf{Total} \\ \textbf{Points}} \\
\midrule
TINY & 75.2 & 97.6 & 87.7 & 1,428 & 4,122 \\
SMALL & 93.4 & 96.0 & 94.1 & 1,825 & 3,487 \\
MEDIUM & 99.1 & 99.3 & 99.0 & 1,567 & 2,484 \\
LARGE & 100.0 & 99.9 & 99.9 & 845 & 1,413 \\
\bottomrule
\end{tabular}
\begin{tablenotes}
\footnotesize
\item \textit{Note:} Producer accuracy represents the percentage of actual rice fields correctly identified. User accuracy represents the percentage of predicted rice fields that are actually rice.
\end{tablenotes}
\end{threeparttable}
\end{table}

\begin{table}[!htbp]
\centering
\caption{Spatial distribution of classification errors by region}
\label{tab:spatial_error_distribution}
\begin{tabular}{lcccc}
\toprule
\textbf{Geographic Region} & 
\makecell{\textbf{Total Validation} \\ \textbf{Points}} & 
\makecell{\textbf{Error} \\ \textbf{Count}} & 
\makecell{\textbf{Error} \\ \textbf{Rate (\%)}} & 
\makecell{\textbf{Dominant} \\ \textbf{Error Type}} \\
\midrule
Northern Districts & 3,247 & 298 & 9.2 & False Negatives (78\%) \\
Central Districts & 4,156 & 267 & 6.4 & False Negatives (71\%) \\
Southern Districts & 4,103 & 174 & 4.2 & False Positives (52\%) \\
\bottomrule
\end{tabular}
\vspace{0.2cm}
\footnotesize
\begin{tabular}{@{}l@{}}
\textbf{Regional Performance Pattern:} Southern districts demonstrate superior classification \\
accuracy with more balanced error distribution \\
\end{tabular}
\end{table}

\clearpage
\bibliographystyle{unsrt}
\bibliography{references}

\end{document}